\documentclass[10pt,letterpaper]{article}
\usepackage[dvipsnames]{xcolor}
\usepackage{multirow}
\usepackage[normalem]{ulem}
\usepackage[top=0.85in,left=2.75in,footskip=0.75in]{geometry}
\usepackage{amsmath,amssymb}
\usepackage{changepage}
\usepackage[utf8x]{inputenc}
\usepackage{textcomp,marvosym}
\usepackage{cite}
\usepackage{nameref,hyperref}

\usepackage[right]{lineno}
\usepackage{microtype}
\usepackage[skip=12pt plus1pt,indent=40pt]{parskip}
\usepackage{array}
\newcolumntype{+}{!{\vrule width 2pt}}
\newlength\savedwidth

\raggedright
\usepackage[aboveskip=1pt,labelfont=bf,labelsep=period,justification=raggedright,singlelinecheck=off]{caption}

\makeatletter
\renewcommand{\@biblabel}[1]{\quad#1.}
\makeatother

\usepackage{lastpage,fancyhdr,graphicx}
\usepackage{epstopdf}
\fancyheadoffset[L]{2.25in}
\fancyfootoffset[L]{2.25in}
\begin{document}
\vspace*{0.2in}
\begin{flushleft}
{\Large
\textbf\newline{A 3-DoF Robotic Platform for the Rehabilitation and Assessment of Reaction Time and Balance Skills of MS Patients} 
}
\newline
\\
Tugce ERSOY\textsuperscript{1},
Elif HOCAOGLU\textsuperscript{2,3*}
\\
\bigskip

\textbf{1} Ozyegin University, Mechanical Engineering, Istanbul, Türkiye
\\
\textbf{2} Living Robotics Laboratory, Istanbul Medipol University, Electrical and Electronics Engineering, Istanbul, Türkiye
\\
\textbf{3} Research Institute for Health Sciences and Technologies (SABITA), Istanbul Medipol University, Istanbul, Türkiye

\bigskip

* ehocaoglu@medipol.edu.tr
\end{flushleft}
\section*{Abstract}
The central nervous system (CNS) exploits anticipatory (APAs) and compensatory (CPAs) postural adjustments to maintain the balance. The postural adjustments comprising stability of the center of mass (CoM) and the pressure distribution of the body influence each other if there is a lack of performance in either of them. Any predictable or sudden perturbation may pave the way for the divergence of CoM from equilibrium and inhomogeneous pressure distribution of the body. Such a situation is often observed in the daily lives of Multiple Sclerosis (MS) patients due to their poor APAs and CPAs and induces their falls. The way of minimizing the risk of falls in neurological patients is by utilizing perturbation-based rehabilitation, as it is efficient in the recovery of the balance disorder. In light of the findings, we present the design, implementation, and experimental evaluation of a novel 3 DoF parallel manipulator to treat the balance disorder of MS. The robotic platform allows angular motion of the ankle based on its anthropomorphic freedom. Moreover, the end-effector endowed with upper and lower platforms is designed to evaluate both the pressure distribution of each foot and the CoM of the body, respectively. Data gathered from the platforms are utilized to both evaluate the performance of the patients and used in high-level control of the robotic platform to regulate the difficulty level of tasks. In this study, kinematic and dynamic analyses of the robot are derived and validated in the simulation environment. Low-level control of the first prototype is also successfully implemented through the PID controller. The capacity of each platform is evaluated with a set of experiments considering the assessment of pressure distribution and CoM of the foot-like objects on the end-effector. The experimental results indicate that such a system well-address the need for balance skill training and assessment through the APAs and CPAs.

\section*{Introduction}
Multiple Sclerosis (MS) is one of the most common chronic neurological diseases in the young adult age group and influences around 2.5 million people in the world. It affects the brain, the central nervous system (CNS), and the spinal cord, thus; it causes balance disorder \cite{Feys_Straudi_2019, Tossavainen_2003, Cattaneo_2014}. In particular, reaching a stable gate pattern takes a longer time, and the patients inevitably do experience recurrent fall events. Moreover, more than seventy-five percent of MS patients experience symptoms of balance disorder\cite{Mehravar_2015}, sixty percent of them encounter at least one fall in 3 months, and more than eighty percent report activity restrictions in their daily lives. Such situations promote a sedentary lifestyle, reduce community and social participation, and lower the living standards of the patients \cite{Tijsma_2017, Aruin_Lee_2015}.

Balance disorder, one of the prominent symptoms due to lesions in the sections of the brain responsible for movement and balance, is common in MS patients \cite{Feys_Straudi_2019, Tossavainen_2003, Cattaneo_2014}. In particular, reaching a stable gate pattern takes a long time for the patients, and they inevitably experience recurrent falls. In the human body, in order to maintain the balance of the whole body, CNS utilizes the anticipatory (feed-forward) and compensatory (feedback) postural adjustments (APAs and CPAs) to control balance. APAs engage and activate the trunk and lower extremity postural muscles before an impending condition/perturbation occurs. It reduces the risk of deterioration in balance by regulating the position of the center of mass (CoM). CPAs are triggered by the sensory control signal and allow the CoM to be repositioned once it has been disrupted. CPAs are activated in both predictable and unpredictable conditions further; when APAs are not able to ensure balance conditions, the body is dependent only on CPAs. These postural adjustments work as coupled, and if one has a deficiency, the other is also affected  \cite{Aruin_Lee_2015, S.Aruin_2016}. MS patients suffer from insufficient balance control due to their poor APAs and CPAs. In the study \cite{Lee_2015}, distortion of their CoM was measured by means of the platform where the patients stood, and the electromyography (EMG) activity levels were evaluated while the patients were catching randomly thrown balls. After regular training, patients were able to keep their CoM in the close vicinity of the equilibrium point and regain their balance ability. Hence, findings in the literature state that perturbation-based rehabilitation programs can improve the balance and decrease the falling rate of MS patients. For instance, the patient may expose to a postural perturbation without any prior information about it \cite {Lee_2015, KANEKAR2015400}.

The other common symptom to diagnose and follow up during the therapy and treatment phase of MS patients is their reaction time against the intentionally provided external stimulation \cite{Saito_2014, Hoang_2016}. Basically, the reaction time is the period that elapses from the start time of the stimulus and the time interval from which the response begins. In particular, MS patients take 1.5 times longer time to stabilize their posture and reach a safer state in a longer period than their healthy pair, which leads to inevitable falling \cite{Cattaneo_2014}. As the longer reaction time is an indication of the loss of balance control, MS disease severity is evaluated and treated as a means of patients' efforts to perturbations. Along these lines, some research groups \cite{Tajali_2018} have focused on the APAs capability of the patients. In this particular, they conducted human-subject experiments on fallers and non-fallers MS patients and observed EMG activity of their leg muscles, pressure distribution, and CoM. The experimental evaluation reveals that MS patients with a history of fall had weaker electrical activity and, as a result, a longer reaction time to regain balance with postural adjustment.

MS patients in the mild-to-moderate disability range are recommended to receive regular rehabilitation therapy four times a week in order to adapt to daily life more easily \cite{Schilling_Lyon_2019}. Despite such an essential need, findings in the literature show that currently, only thirty percent of MS patients benefit from rehabilitation services, and those who could not receive therapy have to use a cane within 20 years and a wheelchair within 30 years from the onset of symptoms \cite{Brown_Kraft_2005}. Due to this reason, demands for therapeutic rehabilitation robots have increased as they can deliver efficient training, and provide objective follow-up assessment and evaluation relative to the conventional rehabilitation techniques \cite{5152604}. As for MS, the research on assistive devices has mainly focused on unilateral or bilateral upper limb dysfunctions since around fifty percent of individuals with MS suffer from impaired hand functions \cite{Lamers_2016, Lamers_Feys_Swinnen_2018}. Some research groups \cite{Feys_2015, Maris_2018} developed a robotic system, I-TRAVLE, for the MS rehabilitation of upper limb dysfunction and forced their impaired arm to perform the tasks defined in the virtual reality (VR) environment. Findings showed that the actuator control ability of the patients was improved and allowed them to lift up their arms to a higher level. Upper limb exoskeletons were also proposed to assist MS patients with paresis \cite{Gijbel_2011}. The study provided evidence that 8-week training with the Armeo Spring improved the patients' upper limb muscle strengths and functional ability.

In the literature, some research groups have also focused on the lower limb rehabilitation of people with MS. For instance, ankle rehabilitation is considered an effective way to improve walking performance, muscle strength \cite{5152604} as well as a range of motion of the ankle of MS patients \cite{Lee_Chen_Ren_Son_Cohen_Sliwa_Zhang_2017}. To address the problem, researchers have mainly utilized two fundamental concepts developed in robotic rehabilitation, i.e., exoskeletons and platform-based robots. Lower limb exoskeletons have been employed for years to treat neurological disorders. In the study \cite{lowerexo_2020}, the exoskeleton together with the assist-as-needed concept was used to explore the efficacy of the MAK exoskeleton for gait rehabilitation of MS patients. Results showed that the MAK exoskeleton improved the gait performance of the patients to some extent; however, loss of stability was observed during the training. Compared to exoskeleton robots, platform-based robots are not wearable and constructed in parallel mechanisms \cite{Fichter_2008}, i.e., a moveable platform connected to the ground with a variety of architectural designs/links, and it transmits forces and executes motions is placed under the patient's foot. Rutgers Ankle, a pneumatic piston-driven 6-Degree of freedom (DoF) parallel manipulator is the pioneer in this category with a VR interface; thus, patients can follow and react in their resistance exercise rehabilitation. It has been substantially documented in the literature that suggested robots can assist to increase ankle muscular strength, gait, and climbing speed \cite{Cioi_2011,Boian_2003,Girone_2000,Deutsch_Boian}. In the study \cite{Gonalves_2014}, the 1-DoF robotic platform (RePAiR) allows for the realization of dorsiflexion and plantar flexion movements in post-stroke patients. The comparison between the patients and the control group revealed that the device availed to increase muscle strength, improved actuator control, sensory-actuator coordination of patients, and accordingly walking patterns. The ARBOT, a 2 DoF robotic platform, \cite{Saglia_2009} was also proposed to train ankle in plantar/dorsiflexion and inversion/eversion directions. Its efficiency was tested with four weeks of rehabilitation and participants significantly improved ankle proprioception. There are also serial platform-based robots, e.g., Optiflex\cite{OptiFlex} and BREVA\cite{Breva}. These 2 DoF robots can perform the dorsiflexion/plantarflexion and inversion/eversion movements, as well. Platforms mentioned above are frequently employed to enhance ankle proprioception and strengthen joint motions. Nevertheless, these systems cannot be utilized to improve postural adjustment and decrease fall rate in MS patients since they only allow for ankle placement in sitting conditions as its end effector area is just one foot large and has a low mass capacity compared to all body weights. However, balance rehabilitation requires training the patient with internal or external perturbations while standing.

Posturography, the expression of postural balance or sway, is evaluated basically by means of platforms with integrated load cells that can provide pressure/force measurement. Such evaluation can be implemented in either static or dynamic means. For the static case, patients are expected to stand on the fixed platform and fulfill the required tasks displayed. While some are doing tasks by means of physical external interactions, like throwing a ball while standing on the platform, the others are performing tasks introduced in the VR environment in order to evaluate their CoM and pressure distribution \cite{Prosperini_Pozzilli_2013}. Some commercially available platforms \cite{Hubbard_Pothier_Hughes_Rutka_2012, Park_Lee_2014, Wii, Tyromotion} exist to assess the performance of the patients under static conditions. They have been put into the services of patients together with the VR environment to give visual feedback and commands and accordingly prolong the training period \cite{Feys_Straudi_2019}. Although static balance platforms are effective in the regulation of pressure distribution \cite{Held_Ferrer_2018}, the findings indicate that rehabilitation of balance disorder in dynamic conditions contributes significantly to the recovery of plasticity \cite{Hubbard_Pothier_Hughes_Rutka_2012}.

Dynamic platforms have been designed to treat balance disorders, like static ones; but what is more, such systems require instantaneous dynamic action of the patients by forcing them to regulate their balance even under perturbations. Various studies have been performed to regulate the balance stability of people. For instance, one of the simplest examples is a wooden balance (wood plate placed on a roller to train the people) \cite{BalanceBoard}. It does not facilitate any sensor therefore, it cannot provide a quantitative assessment.
The ones including load cells are able to measure the pressure of the foot and center of gravity of the patient \cite{Geapro,bobo-balance,CoRehab,Tymo}. Another solution to train the patient has been proposed by GRAIL \cite{GRAIL}, which is a dynamic treadmill working in harmony with the VR environment. Some commercially available robotic platforms \cite{TechnologyLPGMedical, MultitestEquilibre} have been produced to address post-traumatic, orthopedic, or neurological problems by training hip, ankle, and shoulder joints. Depending on the instruction given by the therapist to the robot, the robotic system can intentionally expose patients to an unbalanced state and expect them to strive against the instability. Such systems are able to measure the pressure distribution of each foot and evaluate the balance level of patients by means of the CoM of the body. Human-subject experiments provided evidence that dynamic force platforms help to regain dexterity of the limbs and postural stability compared to the conventional treatment group \cite{Saglia_2019}. The current advanced solution in balance rehabilitation is Hunova\cite{hunova}, which can be utilized in both sitting and standing conditions thus allowing ankle and balance rehabilitation simultaneously. Various clinical studies were performed with this device and showed significant improvements in trunk control, postural balance, and walking speed in different patients group \cite{Cella_2020,Marchesi_2019,Taglione_2018,DeLuca_2020,Naro_2021}. The above-mentioned systems have freedom in 2 axes, roll and pitch axes orientation (plantar/dorsiflexion and inversion/eversion). However, it has been stated that roll, pitch, and z variables are effective in the balance change by mimicking basic real-life activities \cite{Lee_2018}. Despite the benefits of the platforms, there is no study conducted in the literature to investigate pressure distribution training and the ability to control APAs and CPAs, accordingly applied force and the reaction time of the patients, and regulate the control strategy of the platform based on such information. Moreover, the aforementioned robotic platforms were designed as a series of links connected by actuator-actuated joints. Although it is easier to model and do the dynamic analyses of the serial manipulators, preferring parallel manipulators in such an application provides advantages in terms of achieving high load-carrying capability, better dynamic performance, and precise positioning. Considering the efficacy of the parallel manipulator, to the best of our knowledge, there is no such study developed as a parallel manipulator to carry out balance training in the literature. 

In this study, we present the design, simulation, implementation, and experimental evaluation of a 3 DoF parallel manipulator that is designed to rehabilitate the balance dysfunctions of MS patients. The proposed platform is basically composed of three main parts: 3 DoF parallel manipulators, and upper and lower platforms. The 3 DoF parallel manipulator is able to provide both rotational (roll and pitch angles) and linear motion (z-direction) in space thus it can mimic most daily life activities. In addition, it is designed to intentionally destabilize the human body via perturbations in order to force them to regulate and improve their balance since postural adjustment strategies can be improved  by developing compensating mechanisms. The reactions of the patients to maintain their balance while the manipulator is moving are evaluated through the upper and lower platforms. The upper platform is responsible for evaluating the pressure distribution of the feet to follow up on their balance capacity and measuring the reaction time of humans against predictable and unpredictable perturbations. The lower platform is designed to analyze the patients' CoM in real time and points out the severity level of the MS patients. Moreover, the information acquired from the upper and lower platforms is employed as biofeedback to adjust the assistance level required by the assist-as-needed paradigm used in the control architecture of the planar manipulator.

The rest of the manuscript is organized as follows: Section \emph{'Materials and Methods'} introduces the design objectives of the robotic platform, presents kinematic and dynamic analyses of the 3 DoF parallel manipulator and its simulation, while Section \emph{'Results'} evaluates the manipulator's design performance and capability of the upper and lower platforms to analyze the CoM and pressure distribution of the objects in both static and dynamic conditions. Finally, Sections \emph{'Discussion'} and \emph{'Conclusion'} present the discussion and concludes the paper, respectively.

\section*{Materials and Methods}
This section presents the design objectives of the 3 DoF parallel manipulator to be used for the MS patients, kinematic and dynamic analyses of the system, and evaluation of the overall system by means of simulation results.

\subsection*{Design of 3 DoF Robotic Platform}
\label{design_objectives}
The performance requirements of the robotic platform for the rehabilitation of MS patients are categorized based on the presented terminology in \cite{merlet}, of imperative, optimal, primary, secondary, and tertiary requirements.

\begin{figure}[h!]
\begin{center}
\includegraphics[width=12cm]{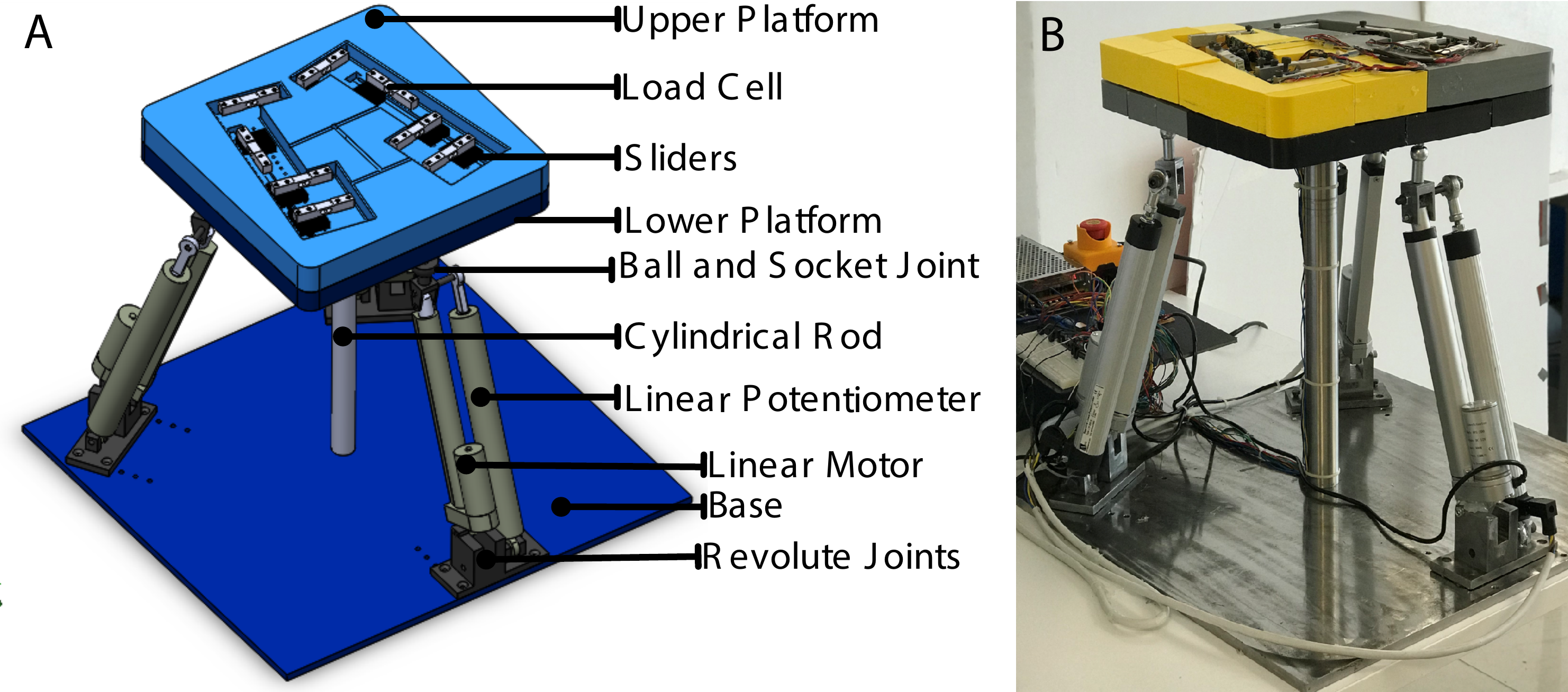}
\end{center}
\caption{3 DoF Robotic Platform \textbf{(A)} In the solid model, the main components of the system are presented \textbf{(B)} The first prototype of the design is presented.}\label{fig:1}
\end{figure}

\begin{figure}[h!]
\begin{center}
\includegraphics[width=12cm]{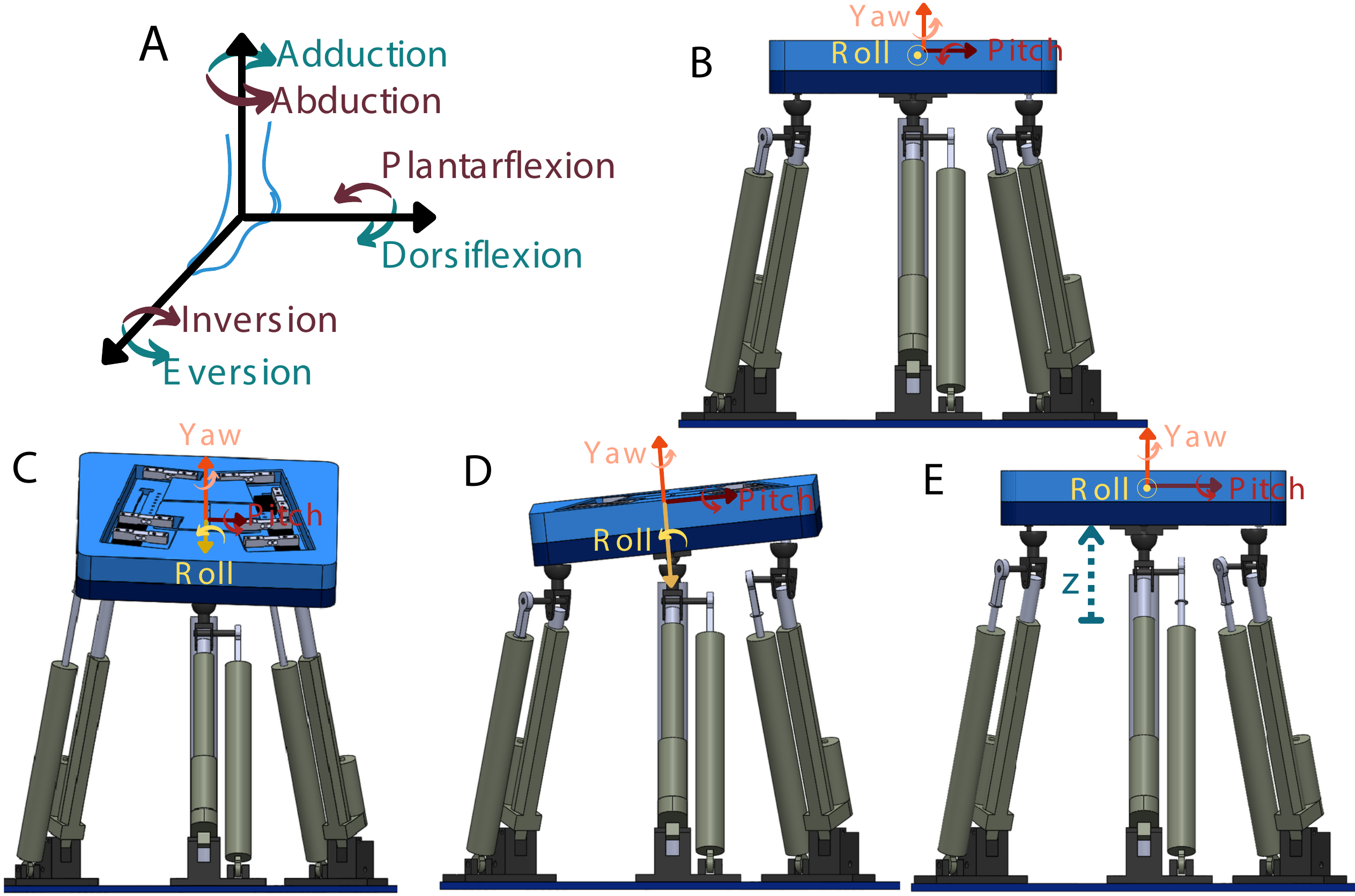}
\end{center}
\caption{\textbf{(A)}~The motions of human ankle joint~\textbf{(B)}~The robotic platform in static condition~\textbf{(C)}~Rotational motion in roll direction~\textbf{(D)}~Rotational motion in pitch direction~\textbf{(E)}~Linear motion in z direction}\label{fig:3}
\end{figure}

Anthropomorphic compliance is an imperative design requirement for the 3 DoF robotic platform (see. Figs \ref{fig:1} and \ref{fig:3}). Enabling patients to exercise within the limit of ankle freedom based on the disorder level is both essential for patients' safety and raising the efficiency of the activity. Besides, doctors/physiotherapists are allowed to adjust the angular constraints of the platform depending on the upper and lower limits of the patients' ability. Accordingly, considering the healthy human, the range of motion of the ankle is identified in three directions with the following ranges: $0-20^o$ dorsiflexion, $0-50^o$ plantarflexion movements, $0-10^o$ adduction, $0-5^o$ abduction, $0-20^o$ eversion and $0-35^o$ inversion \cite{Hasan_Dhingra_2020}. 

The 3 DoF robotic platform can change its orientation in two rotational axes (roll/x and pitch/y angles) and one translation axis (z-direction) in space as shown in Fig \ref{fig:3}, thus; its aim is to improve ankle range of motion, walking performance, muscle strength as well as balance postural adjustments. Although the rotational change in the roll and pitch axes is sufficient for the ankle, the translational movement of the system in the z-axis has an effect on balance, and this elevation in the z-axis is also observed in most of the activities performed in daily life, e.g., taking a step, climbing stairs, repulsion reactions from under the feet. Compared to the devices currently used in balance rehabilitation, it will be possible to enhance postural adjustments with extra translation DoF in the z-axis of the proposed system.

Designing the platform capable of measuring and evaluating the reaction time of the patients during the activities is the optimal performance requirement because the severity of the balance deficit and sensory-based actuator abnormalities may be determined by the reaction time. The proposed design allows physiotherapists to evaluate the balance condition of the patients as well as their cognitive and physical reactions to the physical perturbations applied by the robotic platform itself to compensate for the disturbed balance. Thus, it provides a quantitative evaluation of the robustness of the patients against the physical disturbances while  accompanied by the VR environment to help the patients recover their neuromuscular system with the improvement of the postural response.

The primary design requirement of the platform is identifying the CoM of the body and the pressure distribution of each foot based on the anthropomorphic size of each person. In particular, considering the balance problem of MS patients, the variation of the CoM of their body deviates more dramatically relative to healthy people. Moreover, the unbalanced body gives rise to the unbalanced pressure distribution of the load delivered by the sole of the feet since, in healthy people, almost 50\% of the pressure is applied to the heel, and the other 50\% is applied to the toes, and metatarsals (see Table \ref{table:1}) \cite{Sport_Education_2002}. In other words, the pressure distribution of the feet indicates the intention of patients to balance their body under disturbances and accordingly provides information about the muscular activity during the follow-up. These measurements will be used in the regulation of the assistance level in control architecture to assist the patients as needed. The end-effector of the parallel manipulator is designed as ergonomic and personalized two-fold layers that embody multiple load cells to address the aforementioned issues, i.e., upper and lower platforms. Thus, the proposed platform allows doctors/physiotherapists to evaluate the balance condition of the patients as well as their cognitive and physical reactions to compensate for the disturbed balance while accompanied by the VR environment to help with the improvement of the postural response.

\begin{table}[h!]
\caption{Average foot-pad pressure ratios\cite{Sport_Education_2002}}
\label{table:1}
\begin{tabular}{|c c c c c|}
 \hline
 $N/cm^2$ & Toes & Metatarsals & Middle Foot & Heel of Foot \\
 \hline
 Right & 3.54$\pm$ 3.31 & 28.24$\pm$ 26.18 & 0.97$\pm$ 1.41 & 11.58$\pm$ 9.08 \\
\hline
Left & 3.16$\pm$ 3.64  & 23.85$\pm$ 22.99 & 1.19$\pm$ 1.77 & 12.4$\pm$ 10.26  \\
\hline
\end{tabular}
\end{table}

\begin{figure}[h!]
\begin{center}
\includegraphics[width=12cm]{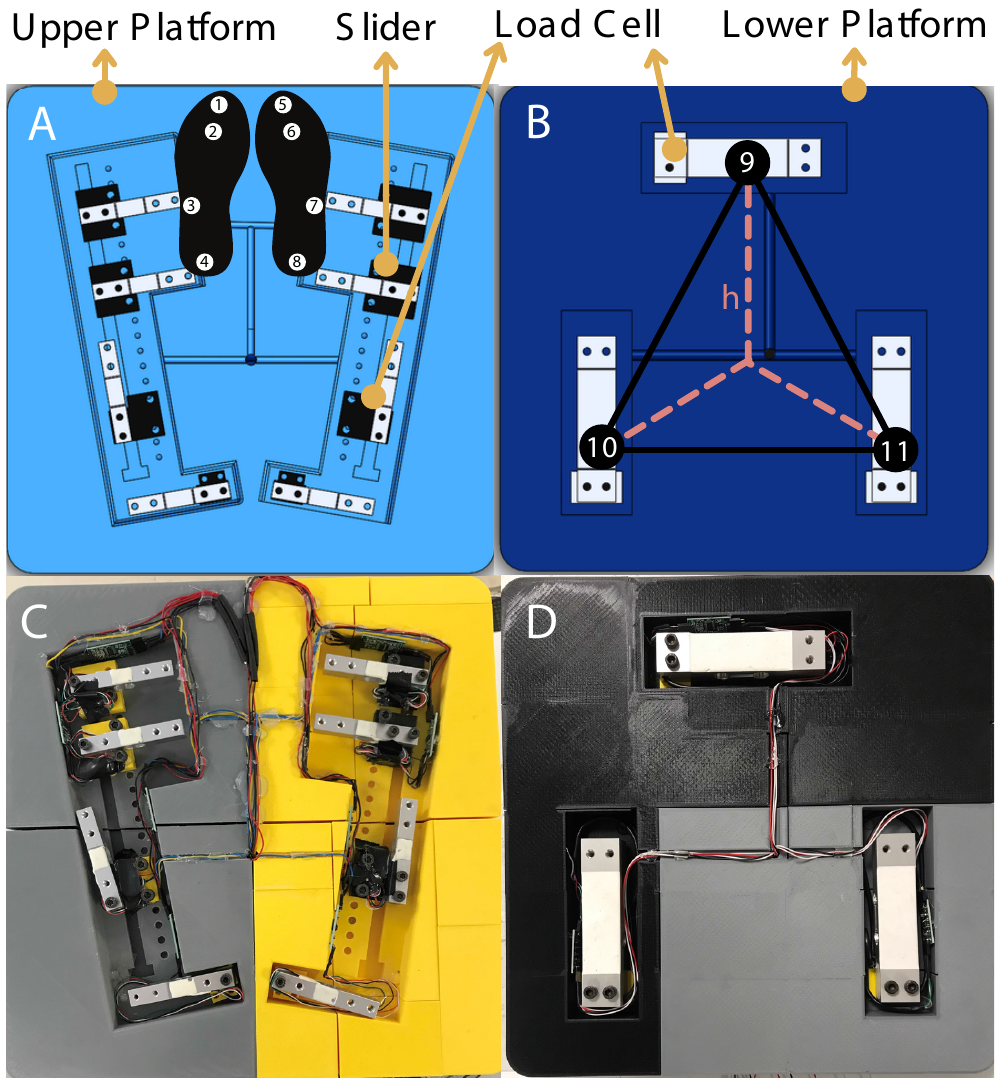}
\end{center}
\caption{Upper and lower platforms~\textbf{(A)}~In the solid model, the components and their positioning on the upper platform are demonstrated~\textbf{(B)}~In the solid model, the components and their positioning on the lower platform are demonstrated~\textbf{(C)}~Top view of the prototype of the upper platform~\textbf{(D)}~Top view of the prototype of the lower platform}
\label{fig:8}
\end{figure}

The upper platform presented as a solid model in Fig~\ref{fig:8}A and as the first prototype in Fig~\ref{fig:8}C responsible for the evaluation of the pressure distribution has two places for each foot. Each part is endowed with four load cells placed under the sole. The proper placement of the load cells is determined based on the effective pressure regions of the sole. In literature, \cite{Sport_Education_2002}, the more densely felt four pressure regions of the foot sole were evaluated, and the pressure values are presented in Table~\ref{table:1}. Load cell mounted on the slider to adjust the position of the sensor and accurately measure the pressure distribution for any foot size. The sliders' distances are determined based on the average anthropometric data of human foot length ~\cite{Gordon_1988,Openshaw_2006,ANTHROPOMETRYBIOMECHANICS} and presented in Table \ref{table:2}. Evaluation of such variation is used to calculate the reaction time of the patients during the rehabilitation. The elapsed time until the patient keeps his/her balance against the perturbation, namely reaction time, is detected and considered as a performance metric for the patient's follow-ups.

 \begin{table}[h!]
\caption{Anthropometric Values\cite{Gordon_1988,Openshaw_2006,ANTHROPOMETRYBIOMECHANICS}}
\label{table:2}
\begin{tabular}{|c c c|}
 \hline
 & Women & Men \\
 \hline
 Height (cm) & 162.94  & 175.58 \\
\hline
 Weight (kg) & 67.12  & 74.74\\
\hline
 Foot Length (cm) & 23.62  & 26.15\\
\hline
 Foot Width (cm) & 9.17  & 10.28\\
\hline
\end{tabular}
\end{table}

The lower platform located just under the upper platform is in charge of the evaluation of the CoM of patients in both static and dynamic conditions. This measurement is utilized to indicate the balance status of the patients as well as give biofeedback to the robotic platform to adjust its assistance level for the patient. The required data to calculate the CoM of the body in the x-y coordinate plane is measured by means of three load cells positioned on the corners of the hidden equilateral triangle placed on the lower platform shown in Figs ~\ref{fig:8}B and \ref{fig:8}D, and it is calculated with Eq~(\ref{eq:77}) and Eq~(\ref{eq:78}). The partial weights of the body distributed over the load cells 9, 10, and 11 are named weight-9, weight-10, and weight-11, respectively, also symbol $h$ represents the distance between each corner and the centroid of the equilateral triangle as depicted in Fig~\ref{fig:8}B.

\begin{eqnarray}
        {COM}_x=\frac{(weight9)h-(h\sin{({30}^o)})weight10-(h\sin{({30}^o)})weight11}{weight9+weight10+weight11}
        \label{eq:77}
\end{eqnarray}

\begin{eqnarray}
        {COM}_y=\frac{(h\cos{({30}^o}) weight10-(h \cos{({30}^o)}) weight11}{weight10+weight11}
        \label{eq:78}\end{eqnarray}

The secondary requirement for the overall system design is to guarantee the safety of the patients standing on the platform under dynamic conditions. The 3 DoF parallel manipulator is designed to be able to carry the average human weight, around 687 N, as well as the weight of the end-effector (lower and upper platforms), around 120 N. That being said, each linear actuator has a 900 N force capacity to resist more than the maximum load and is positioned under the end-effector with equal distance as seen in Fig~\ref{fig:1}. In order to ensure the safety of the system and its performance, the kinematic and dynamic analyses of the platform have been implemented both in simulation and real environments with the proof of concept design.

\begin{figure}[h!]
\begin{center}
\includegraphics[width=12cm]{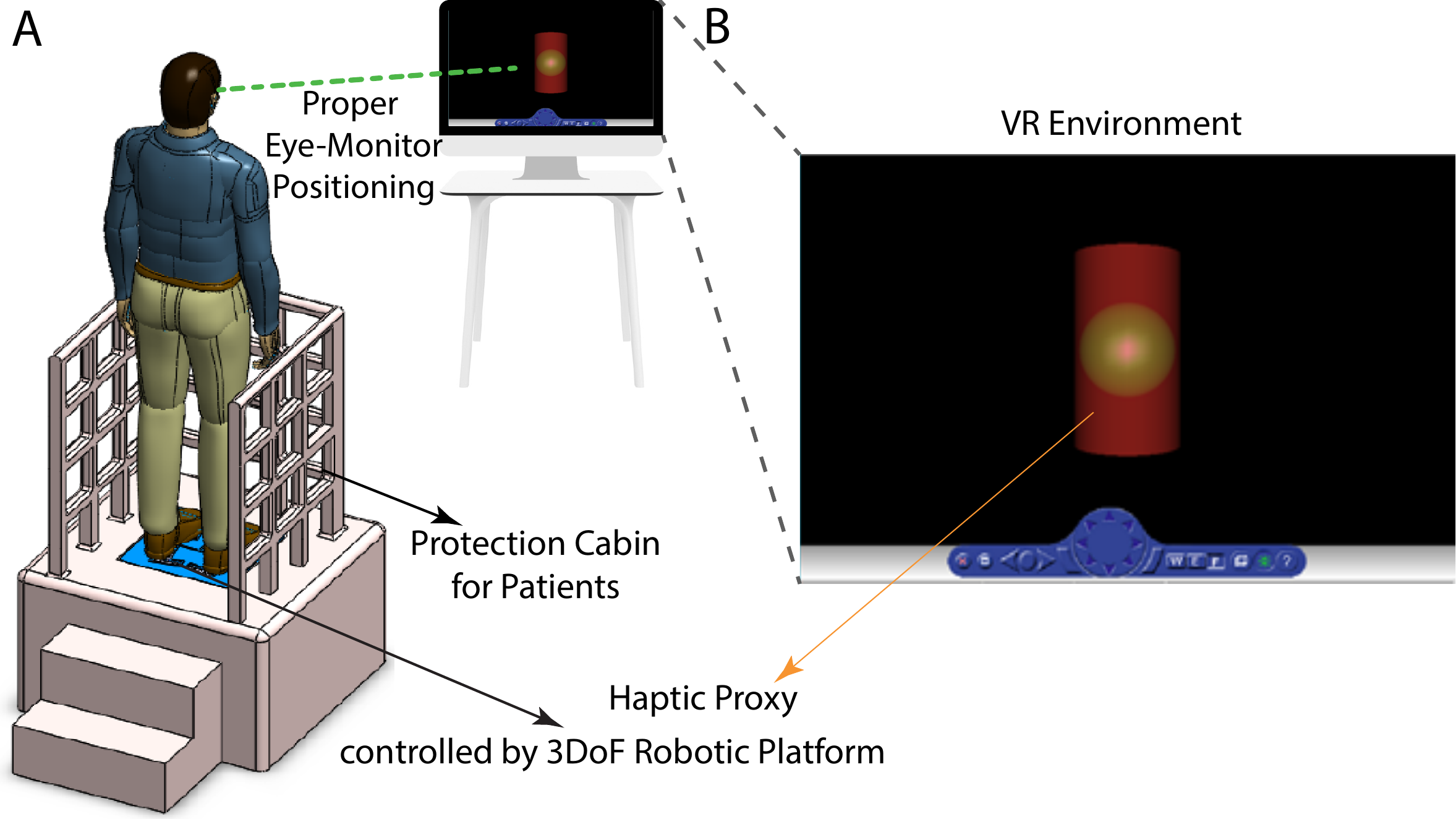}
\end{center}
\caption{\textbf{(A)}Illustration of the ready-to-use system configuration~\textbf{(B)}Virtual reality environment}\label{fig:2}
\end{figure}

The tertiary requirement of the system's design criteria is increasing the efficiency of the training to get higher benefits from rehabilitation. To address the need of the criterion, the physical system is accompanied by the VR environment in such a way that the patients are both motivated by means of the tasks defined in the VR and conducted in a correct manner to do activities properly. The VR environment as depicted in Fig~\ref{fig:2}a is placed in front of the patients to satisfy body-eye coordination properly. The task in the VR environment is analogously designed with the activities forced by the robotic platform. For instance, the angular deviation of the end-effector is matched with the greenish ball, a.k.a. haptic proxy, presented in Fig~\ref{fig:2}b. In this example, it is expected from the patient to keep the green ball in the oscillatory red cylinder which moves up and down with random frequencies by controlling the angular position of the end-effector. While the red cylinder is perturbed, the patient physically feels the same action by the robotic platform and responds to it to control the haptic proxy. During such activities, the system is able to measure the reaction time of the patient's attempt to the physical perturbations. Moreover, measured data allow us to quantitatively evaluate the patient's APAs and CPAs in his/her intention phase to maintain the balance of the posture.

\subsection*{Kinematic and Dynamic Analyses of 3 DoF Parallel Manipulator}

In this Section, we detail the kinematic analysis of the 3 DoF parallel manipulator constructing the vector loop equations and the system's dynamic analysis based on the Euler-Lagrange method.

\begin{figure}[h!]
\begin{center}
\includegraphics[width=12cm]{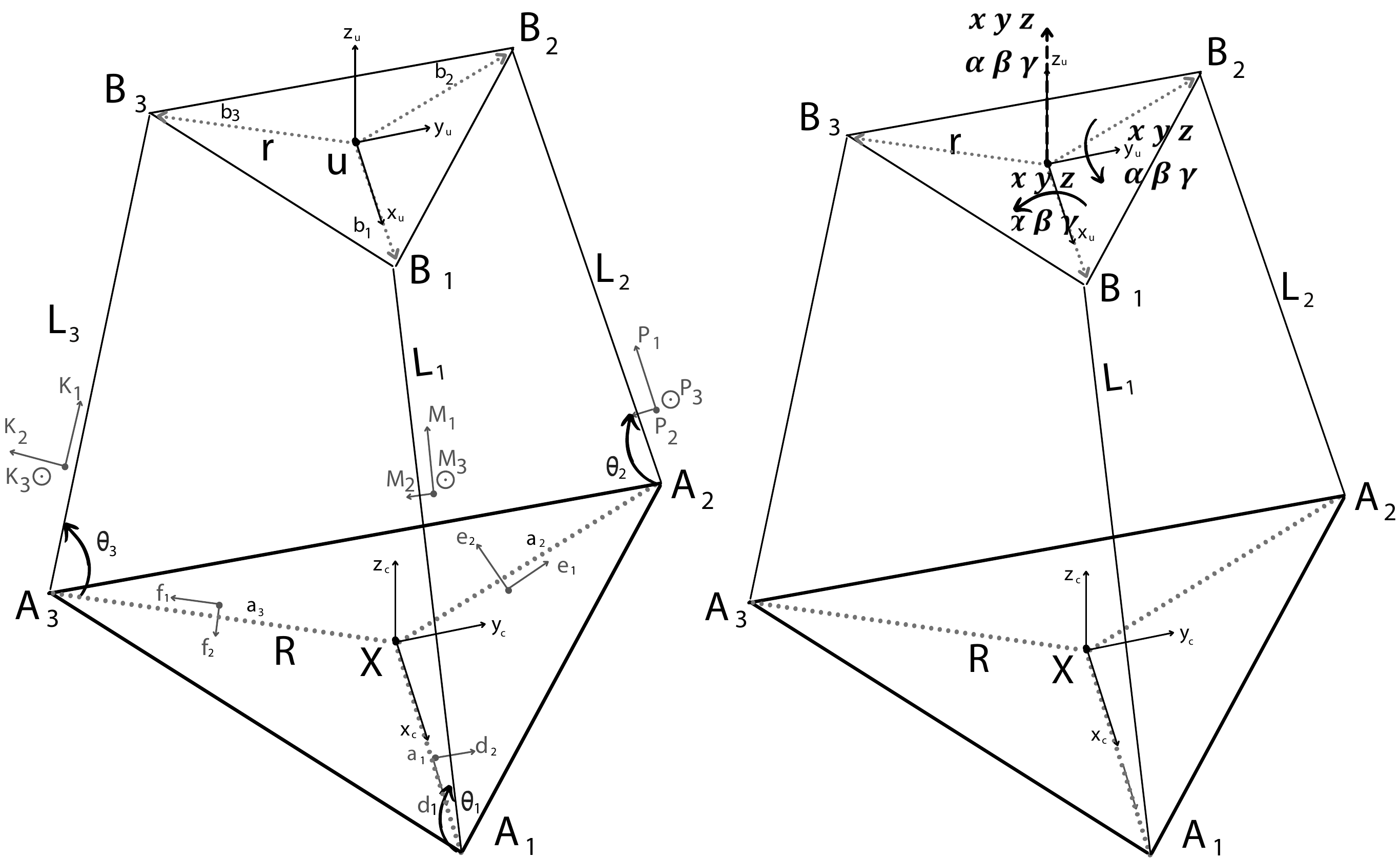}
\end{center}
\caption{Schematic representation of 3 DoF parallel manipulator}\label{fig:4}
\end{figure}

The architecture of the 3-DoF parallel manipulator is shown in Fig. 1. 
It comprises an end effector (a two-fold layer, e.g., upper and lower platform), a fixed base, and three supporting limbs/linear actuators with an exact kinematic structure. The cylindrical rod has a fixed length and does not play any role in dynamic conditions. Its only role is to support the system in a static condition, that is when the system does not move. Accordingly, the rod element is not taken into account in the kinematic and dynamic derivations of the platform. 
Fig~\ref{fig:4} presents the schematic representation of the 3 DoF parallel manipulator where the bold triangular lines describe the base and end effector. The corners of the upper triangle typify the ball and socket, whereas the corners of the lower triangle typify the revolute joint. The main reason for selecting ball and socket and revolute type joints in such a system is to ensure that the motion of each actuator does not affect the other.

The base is attached to the end effector by three identical PRS linkages, i.e., a sequence P (Prismatic) joint, an R (Revolute) joint, and an S (Spherical) joint, where a linear actuator actuates the P joint and others are passive joints. The end effector changes its orientation on 6-axis, but only 3 of them are controllable, and others are solved with the constraint equation, which is explained in Section of \emph{'Kinematic Analysis of the Manipulator'}. Two DoF of the manipulator, regulated by roll$(\alpha)$-pitch$(\beta)$ angles, are assigned for the orientations in two directions, while one DoF of the manipulator is allocated for the translation in the z-direction. 

The mobility criterion of the proposed 3-DoF parallel manipulator is proven with Eq. \ref{eq:hunt}  \cite{Carretero_2000}.

\begin{eqnarray}
 DoF=6(n-g-1)+ \sum_{i=1}^{g}f_i\label{eq:hunt}
\end{eqnarray}

Eq. \ref{eq:hunt} calculates the DoF of the system. \emph{'n', 'g', and '$f_i$'} represent the number of bodies, joints, and degree of freedom of each joint, respectively. In the proposed platform, \emph{'n'} is equal to 8 (3 linear actuators, 3 passive joints, 1 end effector (upper and lower platforms are mounted each other), and fixed based), \emph{'g'} is equal to 9 (3 ball and socket, 3 prismatic, and 3 revolute), \emph{'$f_i$'} for ball and socket joint is 3, whereas for prismatic and revolute joints is equal to 1. Eq. \ref{eq:hunt} results in 3, Eq. \ref{eq:hunt2}, which also proves that the proposed robotic platform has 3-DoF.
\begin{eqnarray}
 DoF=6(8-9-1)+ (3(1) + 3(1) + 3(3))= 3\label{eq:hunt2}
\end{eqnarray}
\subsubsection*{Kinematic Analysis of the Manipulator}
\label{Kinematics}
The proposed solution for the kinematic analysis utilizes three vector loop equations to construct the geometric relations among the frames. As depicted in Fig~\ref{fig:4}, the end-effector and the base frame are geometrically expressed as a means of two equilateral triangles connected by frames of three linear actuators. Here, the symbol $r$ represents the distance between the corners of the triangles and the center of the platform, also symbol $\theta_i$ defines the positioning of the actuators on the base frame and is fixed to $70^{\circ}$.
     
Symbols $X$ and $u$ are the centers of the base platform and upper platform (end-effector) as depicted in Fig~\ref{fig:4}, respectively. A fixed coordinate frame \textbf{(X)} is established on the base platform and the coordinates of the frame are named $x_c$, $y_c$, and $z_c$.

Coordinates $A_1$, $A_2$, and $A_3$ of the corners of the base frame are interpreted with respect to \textbf{X} frame by means of the position vectors in expression~\ref{eq:01}.

	\begin{eqnarray}
	A_1= \begin{bmatrix} R\\ 0\\0\\ \end{bmatrix},A_2= \begin{bmatrix} -1/2R\\\sqrt{3}/2R\\ 0\\\end{bmatrix}, A_3= \begin{bmatrix}-1/2R\\-\sqrt{3}/2R\\0\\\end{bmatrix}\label{eq:01}
	\end{eqnarray}

    The coordinates of the corners of upper platform depicted in Fig~\ref{fig:3}, $b_1$, $b_2$, and $b_3$ with respect to \textbf{u} frame are described in~(\ref{eq:02}).
	\begin{eqnarray} b_1= \begin{bmatrix} r\\ 0\\0\\ \end{bmatrix},b_2= \begin{bmatrix} -1/2r\\\sqrt{3}/2r\\ 0\\\end{bmatrix},   b_3= \begin{bmatrix}-1/2r\\-\sqrt{3}/2r\\0\\\end{bmatrix}\label{eq:02}\end{eqnarray}

     The frame \textbf{u} can be expressed with respect to the \textbf{X} frame by using Euler Angle Representation. In Matrix~(\ref{eq:03}), symbols $(\alpha)$, $(\beta)$ and $(\gamma)$ represent roll (x-axis rotation), pitch (y-axis rotation), and yaw (z-axis rotation) angles, respectively. Accordingly, the coordinates, $b_1$, $b_2$, and $b_3$ stated with respect to the \textbf{X} frame by means of the Euler Angle transformation,~\textit{T} is expressed in matrix notation as:

    \begin{eqnarray}
    T= \begin{bmatrix} C_\alpha C_\beta & C_\alpha S_\beta S_\gamma-S_\alpha C_\gamma & C_\alpha S_\beta C_\gamma+S_\alpha S_\gamma & X_U  \\ S_\alpha C_\beta & S_\alpha S_\beta S_\gamma+C_\alpha C_\gamma & S_\alpha S_\beta C_\gamma-C_\alpha S_\gamma & Y_U \\ -S_\beta & C_\beta S_\gamma & C_\beta C_\gamma & Z_U \\
    0 &0 &0 &1 \\\end{bmatrix}\label{eq:03}
    \end{eqnarray}

    The position vectors $b_1$, $b_2$, and $b_3$ with respect to fixed frame~(\textbf{X}) are typified with symbols $B_1$, $B_2$, and $B_3$, respectively. Here, \textit{C} and \textit{S} symbolize the cosine and sine functions, respectively.

    \begin{eqnarray}
    \begin{bmatrix} B_i\\ 1\\\end{bmatrix}=\begin{bmatrix}T\\\end{bmatrix}\begin{bmatrix} b_i\\1\end{bmatrix}\label{eq:04}
    \end{eqnarray}

    \begin{eqnarray}
    B_1= \begin{bmatrix} r C_\alpha C_\beta+X_U\\ r S_\alpha C_\beta+Y_U\\-r S_\beta+Z_U\end{bmatrix}\label{eq:05}
    \end{eqnarray}

     \begin{eqnarray}
      B_2= \begin{bmatrix}  \frac{-r C_\alpha C_\beta+\sqrt{3} r (C_\alpha S_\beta S_\gamma-S_\alpha C_\gamma)}{2}+X_U\\\frac{-r S_\alpha C_\beta+\sqrt{3} r (S_\alpha S_\beta S_\gamma+C_\alpha C_\gamma)}{2}+Y_U\\\frac{r S_\beta+\sqrt{3} r C_\alpha S_\gamma}{2}+Z_U\end{bmatrix}\label{eq:06}
    \end{eqnarray}

    \begin{eqnarray}
    B_3=\begin{bmatrix}  \frac{-r C_\alpha C_\beta-\sqrt{3} r (C_\alpha S_\beta S_\gamma-S_\alpha C_\gamma)}{2}+X_U\\\frac{-r S_\alpha C_\beta-\sqrt{3} r (S_\alpha S_\beta S_\gamma+C_\alpha C_\gamma)}{2}+Y_U\\\frac{r S_\beta-\sqrt{3} r C_\alpha S_\gamma}{2}+Z_U\end{bmatrix}\label{eq:07}
    \end{eqnarray}

\begin{figure}[h!]
\begin{center}
\includegraphics[width=12cm]{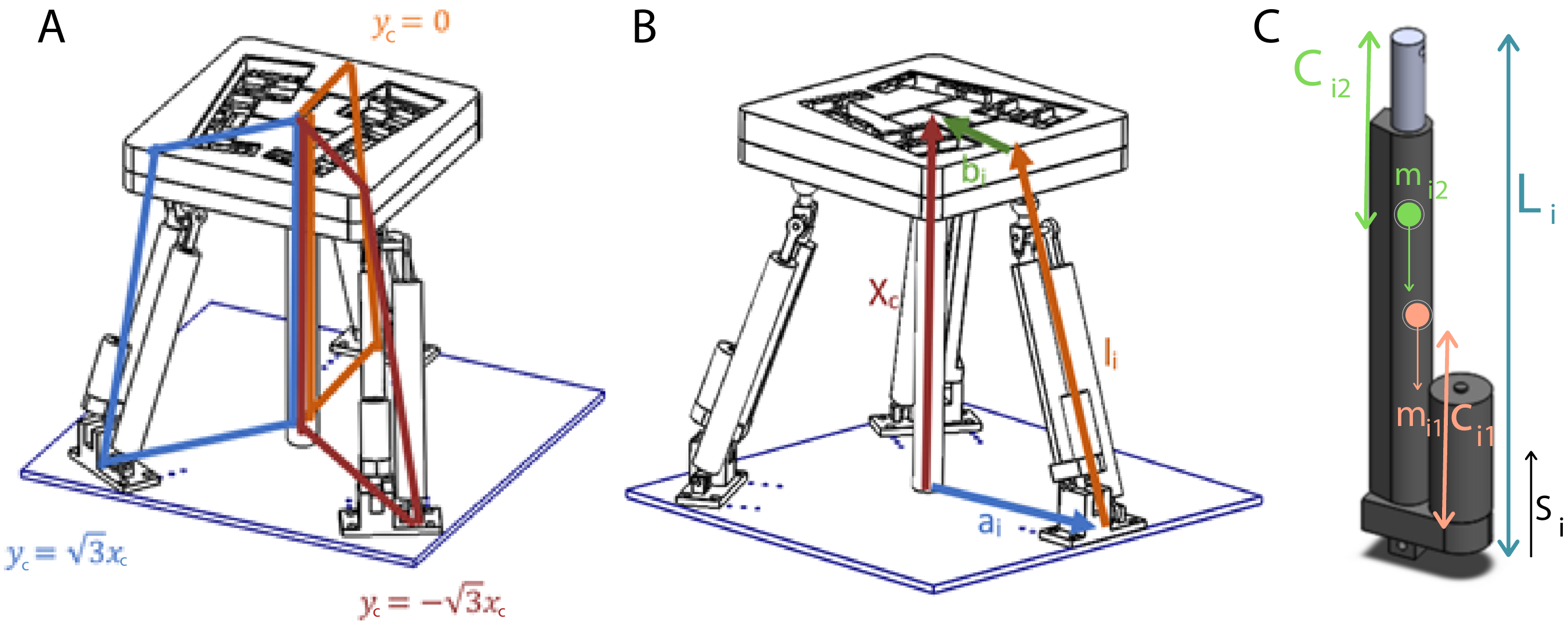}
\end{center}
\caption{\textbf{(A)} Geometric constraints~\textbf{(B)} Vector loops for the kinematic and dynamic analyses ~\textbf{(C)} Linear actuator link variables}\label{fig:5}
\end{figure}

    Actuators' positions on \textbf{X} frame identify the constraints of the kinematics as illustrated in Fig~\ref{fig:5}A. The constraint equations are stated for each actuator's static body as $y_c=0$ (first actuator's constrain plane), $y_c=\sqrt3\ x_c$ (second actuator's constrain plane) and $y_c=-\sqrt3\ x_c$ (third actuator's constrain plane). The constraint equations are derived as in Eqs~(\ref{eq:08}-\ref{eq:10}).
    
    \begin{eqnarray}
     r S_\alpha C_\beta+Y_U=0\label{eq:08}\end{eqnarray}

   \begin{eqnarray}
      \frac{\sqrt3 r(S_\alpha S_\beta S_\gamma + C_\alpha C_\gamma)-r S_\alpha C_\beta +2Y_U}{-2\sqrt3}=\frac{\sqrt3 r(C_\alpha S_\beta S_\gamma - S_\alpha C_\gamma)-r C_\alpha C_\beta + 2X_U}{2}\label{eq:09}
    \end{eqnarray}

     \begin{eqnarray}
    \frac{2Y_U-r S_\alpha C_\beta-\sqrt3 r(S_\alpha S_\beta S_\gamma+C_\alpha C_\gamma)}{2\sqrt3}= \frac{2X_U-r C_\alpha C_\beta-\sqrt3 r(C_\alpha S_\beta S_\gamma-S_\alpha C_\gamma)}{2}\label{eq:10}
    \end{eqnarray}

    Eq~(\ref{eq:11}) and Eq~(\ref{eq:12}) are derived to represent the constraint equations in \textbf{u} frame by adding and subtracting Eq~(\ref{eq:09}) and Eq~(\ref{eq:10}).

    \begin{eqnarray}
      {S_\alpha C_\beta=C}_\alpha S_\beta S_\gamma-S_\alpha C_\gamma\label{eq:11}
    \end{eqnarray}

     \begin{eqnarray}
      X_U=\frac{r(C_\alpha{C}_\beta-S_\alpha{S}_\beta S_\gamma-C_\alpha{C}_\gamma)}{2}
      \label{eq:12}
      \end{eqnarray}

      Eq~(\ref{eq:11}) satisfies the equality for any integer of \textit{n}; however considering the physical limitation of the system requirement, the summation of~$\alpha$ and~$\gamma$ cannot exceed zero as expressed in Eq~(\ref{eq:13}).

    The constraint Eqs(\ref{eq:13}-\ref{eq:15}) are rearranged to diminish the number of unknowns from five (x, y, roll$(\alpha)$, pitch$(\beta)$ and yaw$(\gamma)$) to two (roll$(\alpha)$ and  pitch$(\beta)$). The relations among the x, y, z, roll$(\alpha)$, pitch$(\beta)$ and yaw$(\gamma)$ angles expressed in Eqs~(\ref{eq:11}-\ref{eq:12}) are utilized to determine the constraint Eqs~(\ref{eq:13}-\ref{eq:15}) to characterize the upper platform by $X_U$ and $Y_U$ depicted in Fig~\ref{fig:5}A.

      \begin{eqnarray}
      \alpha+\gamma=n\pi,~~where~~n=0\label{eq:13}
      \end{eqnarray}
      \begin{eqnarray}
       X_U=\frac{-r(1-C_\beta)C_{2\alpha}}{2}\label{eq:14}
      \end{eqnarray}
      \begin{eqnarray}
       Y_U=\frac{r(1-C_\beta)S_{2\alpha}}{2}\label{eq:15}
      \end{eqnarray}

     The connection between the upper and lower platforms is carried out via three linear actuators. Considering the kinematic point of view, the vector-based representation of the actuators' positions in the space is defined by their lengths, $l_i$, where $i=1,2,3$. The Pythagorean theorem allows us to establish the fundamental relation in Euclidean geometry among the three sides of a right triangle as depicted in Fig~\ref{fig:5}B and the formula to express the length of each actuator is presented in Eq~(\ref{eq:16}).
      	
      \begin{eqnarray}
       l_i=\sqrt{(B_{ix}-A_{ix})^2+(B_{iy}-A_{iy})^2+(B_{iz}-A_{iz})^2}\label{eq:16}
      \end{eqnarray}

     Accordingly, the length of each actuator is expressed as a function of three independent variables of roll$(\alpha)$, pitch$(\beta)$, and $Z_{U}$.

      \begin{eqnarray}
      l_1 =\sqrt{(r C_\alpha C_\beta+X_U-R)^2+(r S_\alpha C_\beta + Y_U)^2+({-r S}_+{\ Z}_U)^2}\label{eq:17}\end{eqnarray}
      \begin{eqnarray}
       l_2=\sqrt{\begin{matrix}(\frac{\sqrt3r(C_\alpha S_\beta S_\gamma-S_\alpha C_\gamma)-r C_\alpha C_\beta+2X_U+R}{2})^2..
      \\+(\frac{2Y_U-r S_\alpha C_\beta+\sqrt3 r(S_\alpha S_\beta S_\gamma+C_\alpha C_\gamma)-\sqrt3R}{2})^2+(\frac{r S_\beta +\sqrt3 r(C_\beta S_\gamma)+2Z_U}{2})^2\end{matrix}}\label{eq:18}
      \end{eqnarray}
      
      \begin{eqnarray}
       l_3=\sqrt{\begin{matrix}(\frac{2X_U+R-rC_\alpha C_\beta-\sqrt3r (C_\alpha S_\beta S_\gamma-S_\alpha C_\gamma)}{2})^2... \\+(\frac{2Y_U-rS_\alpha C_\beta-\sqrt3r (S_\alpha S_\beta S_\gamma+C_\alpha C_\gamma)+\sqrt3R}{2})^2+(\frac{rS_\beta+\sqrt3 r(C_\beta S_\gamma+2Z_U)}{2})^2\end{matrix}}\label{eq:19}
      \end{eqnarray}

  The kinematic architecture of the parallel manipulator is constructed based on the closed-loop kinematic chains as illustrated in Fig~\ref{fig:5}A. In order to solve the forward kinematic problem for the manipulator based on the position variation of each actuator as a function of time, three vector loop equations are derived by summing up the vectors presented in Fig~\ref{fig:5}B and expressed in Eq~(\ref{eq:20}). Similarly, the inverse kinematics of the manipulator is formulated by rearranging the vector loop Eqs~ (\ref{eq:21}-\ref{eq:23}) to determine the instantaneous actuator position required for the end effector to reach the desired position.

     \begin{eqnarray}
      X_{Ui}|_{n1}+Y_{Ui}|_{n2}+Z_{Ui}|_{n3}-R|R_0^{Ai}(1:3,1)-l_i|R_0^{Bi}(1:3,1)=0\label{eq:20}
         \end{eqnarray}
         \begin{eqnarray}
      X_{U1}|_{n1}+Y_{U1}|_{n2}+Z_{U1}|_{n3}-R|R_0^{A1}(1:3,1)-l_1|R_0^{B1}(1:3,1)=0\label{eq:21}
         \end{eqnarray}
         \begin{eqnarray}
      X_{U2}|_{n1}+Y_{U2}|_{n2}+Z_{U2}|_{n3}-R|R_0^{A2}(1:3,1)-l_2|R_0^{B2}(1:3,1)=0\label{eq:22}
         \end{eqnarray}
         \begin{eqnarray}
      X_{U3}|_{n1}+Y_{U3}|_{n2}+Z_{U3}|_{n3}-R|R_0^{A3}(1:3,1-l_3|R_0^{B3}(1:3,1)=0\label{eq:23}
         \end{eqnarray}
         where rotational matrices;

          \begin{eqnarray}
       R_0^{A1}=\left[\begin{matrix}1&0&0\\0&1&0\\0&0&1\\\end{matrix}\right],R_0^{B1}=R_0^{A1} R_{A1}^{B1}=\left[\begin{matrix}-\cos{\theta_1}&0&\sin{\theta_1}\\0&1&0\\\sin{\theta_1}&0&\cos{\theta_1}\\\end{matrix}\right]\label{eq:24}
        \end{eqnarray}

         \begin{eqnarray}
       R_0^{A2}=\left[\begin{matrix}-\cos{60^o}&sin60^o&0\\sin60^o&-cos60^o&0\\0&0&1\\\end{matrix}\right],\ \ \ R_0^{B2}=R_0^{A2} R_{A2}^{B2}=\left[\begin{matrix}\frac{\cos{\theta_2}}{2}&\frac{\sqrt3}{2}&-\frac{\sin{\theta_2}}{2}\\-\frac{\sqrt3}{2}\cos{\theta_2}&\frac{-1}{2}&\frac{\sqrt3}{2}\sin{\theta_2}\\\sin{\theta_2}&0&\cos{\theta_2}\\\end{matrix}\right]\label{eq:25}
         \end{eqnarray}

         \begin{eqnarray}
      R_0^{A3}=\left[\begin{matrix}-cos{60^o}&sin60^o&0\\-sin60^o&cos60^o&0\\0&0&1\\\end{matrix}\right], R_0^{B3}=R_0^{A3} R_{A3}^{B3}=\left[\begin{matrix}\frac{\cos{\theta_3}}{2}&\frac{\sqrt3}{2}&-\frac{\sin{\theta_3}}{2}\\\frac{-\sqrt3}{2}\cos{\theta_3}&\frac{-1}{2}&\frac{-\sqrt3}{2}\sin{\theta_3}\\\sin{\theta_3}&0&\cos{\theta_3}\\\end{matrix}\right]\label{eq:26}
         \end{eqnarray}

      After the rearrangement of the Eqs~(\ref{eq:21}-\ref{eq:26}), the coordinate of each linear actuator in \textbf{u} frame is expressed by Eqs~(\ref{eq:27}-\ref{eq:29}).

      \begin{eqnarray}
       X_{U1}=R-l_1\cos{(\theta_1)}, Y_{U1}=0, Z_{U1}=l_1\sin{(\theta_1)}\label{eq:27}
      \end{eqnarray}
       \begin{eqnarray}
       X_{U2}=\frac{-R+l_2\cos{(\theta_2)}}{2}, Y_{U2}=\frac{\sqrt{3(}R-l_2\cos{(\theta_2)})}{2}, Z_{U2}=l_2\sin{(\theta_2)}\label{eq:28}
      \end{eqnarray}
      \begin{eqnarray}
       X_{U3}=\frac{-R+l_3\cos{(\theta_3)}}{2}, Y_{U3}=\frac{\sqrt{3(}-R+l_3\cos{(\theta_3)})}{2}, Z_{U3}=l_3\sin{(\theta_3)}\label{eq:29}
      \end{eqnarray}

      The positioning of the actuators on the end-effector forming the corners of the triangular shape depicted in Fig~\ref{fig:4} enables us to calculate the coordinate of the centroid of the triangle, namely the end-effector ($X_c$, $Y_c$, $Z_c$) in Eq~\ref{eq:30} by averaging the well-defined coordinates, $X_U$, $Y_U$, $Z_U$.

      \begin{eqnarray}
   X_c=\frac{\sum_{i=1}^{3}X_{Ui}}{3}, Y_c=\frac{\sum_{i=1}^{3}Y_{Ui}}{3} ,Z_c=\frac{\sum_{i=1}^{3}Z_{Ui}}{3}\label{eq:30}
      \end{eqnarray}

\subsubsection*{Dynamic Analysis of the Manipulator}
\label{sec:manip_dyn}
The 3-DoF parallel manipulator considered in this work is equipped with three linear actuators mounted between the end-effector and the base of the robot. The end-effector is able to rotate in roll $(\alpha)$ and pitch $(\beta)$ angles and moves in the z-axis as represented in Fig~\ref{fig:3}. The dynamic model of a 3-DoF parallel manipulator in joint space is derived utilizing the Euler-Lagrange method. Accordingly, the solutions extract the required forces/torques that maintain the system to follow the reference trajectory.

The general form of the dynamic equation is,
      \begin{eqnarray}
        M(X)\ddot{X}+C(X,\dot{X})+G=F\label{eq:31}
      \end{eqnarray}
where ${X}$ symbolizes the generalized coordinate vector of the manipulator, ${M}$ represents the mass vector, ${C}$ is the centrifugal and Coriolis matrix, ${G}$ is the gravity vector, and ${F}$ is a vectorial representation of the required forces to actuating each linear motor in task space.

The kinematics of the platform is structured based on the unit vector $(\widehat{s_i})$ along the actuator direction utilizing the loop equations discussed in the kinematics analysis of the manipulator.

To find the dynamic equation, the closed-loop equation, and link lengths are redefined by using unit vector $(\widehat{s_i})$ along the actuator direction. $\widehat{s_1}$, $\widehat{s_2}$, and $\widehat{s_3}$ are unit vectors are in the same direction as axes K1, M1, and P1 shown in Fig \ref{fig:3}, respectively
       \begin{eqnarray}
          X_c=A_i+l_i\widehat{s_i}-B_i\label{eq:32}
      \end{eqnarray}
      As found in inverse kinematics, the length of each actuator found as
        \begin{eqnarray}
          l_i=\mid\mid X_c+B_i-A_i\mid\mid_2\label{eq:33}
      \end{eqnarray}
      To express the length of the actuators in the coordinate frame, unit vectors are defined as,
      \begin{eqnarray}
          \widehat{s_i}=\frac{X_c+B_i-A_i}{l_i}\label{eq:34}
      \end{eqnarray}

      An intermediate variable, $\delta_i$, is assigned as the position of the $B_i$ points represented in Fig~\ref{fig:4} and derived as a means of vector summations in two different vector loop structures as shown in Fig~\ref{fig:5}B,

      \begin{eqnarray}
          \delta_i=X_c+B_i\label{eq:35}
      \end{eqnarray}
       \begin{eqnarray}
         \delta_i=A_i+l_i\widehat{s_i}\label{eq:36}
      \end{eqnarray}
      By differentiation Eq~ (\ref{eq:35}) and Eq~(\ref{eq:36}), velocities of intermediate variable can be found as,
       \begin{eqnarray}
         \dot{\delta_i}=\dot{X_c}+w\times B_i\label{eq:37}
      \end{eqnarray}
       \begin{eqnarray}
         \dot{\delta_i}=\dot{l_i}\widehat{s_i}+l_i w_i\times\widehat{s_i}\label{eq:38}
      \end{eqnarray}
      where $w_i$ and $\dot{l_i}$ represent the angular velocity and length rate of each actuator, respectively. Symbol $w$ is the angular velocity of the end effector. Again, by differentiation Eq~(\ref{eq:37}) and Eq~(\ref{eq:38}), acceleration of intermediate variable can be found as,
       \begin{eqnarray}
         \ddot{\delta_i\ }=\ddot{X_c}+w\times B_i\ +w\times(w\times B_i)
        \label{eq:39}
       \end{eqnarray}
         \begin{eqnarray}
         \ddot{\delta_i}=\ddot{l_i}\widehat{s_i}+l_i w_i\times\widehat{s_i}+\dot{l_i}w_i\times \widehat{s_i}+l_i\dot{w_i}\times \widehat{s_i}+{l_i w}_i\times(w_i\times\widehat{s_i}\ )\label{eq:40}
       \end{eqnarray}
       Then by applying dot product between Eq~(\ref{eq:37}) and $\widehat{s_i}$, angular and linear velocity of the intermediate variable can express with the position of the actuators,
       \begin{eqnarray}
        \dot{l_i}=\dot{\delta_i}\widehat{s_i}=[\dot{X_c}+w\times B_i]^T.s_i \label{eq:41}
       \end{eqnarray}
       Applying cross product procedure in Eq~(\ref{eq:37}) and Eq~(\ref{eq:38}) to $\widehat{s_i}$, angular velocity of each actuator yield,
       \begin{eqnarray}
       w_i=\frac{\widehat{s_i}\times\dot{\delta_i}}{l_i}=\frac{\widehat{s_i}\times(\dot{X_c}+w\times B_i)}{l_i}\label{eq:42}\end{eqnarray}
        In order to make calculations easier, dynamic analysis of the actuators and moving platforms were done separately and later combined with the Jacobian matrix.

\begin{enumerate}
\item{\textbf{Dynamic Analysis of Each Actuation Unit}}
\label{item:act_dyn}

The dynamic equation in Eq~(\ref{eq:31}) stated in task space is reformed and expressed as Eq~(\ref{eq:43}) to find the dynamic equation of each actuator in joint space.

        \begin{eqnarray}
          M_i\ddot{\delta_i}+C_i\dot{\delta_i}+G_i=F_{i}\label{eq:43}\end{eqnarray}
 where $M_i$, $C_i$, $G_i$, and $F_i$ represent mass matrix, centrifugal and Coriolis matrix, gravity vector of the actuators, and generated force by each actuator, respectively.

The assembly of the linear actuator is composed of two main parts as represented Fig~\ref{fig:5}C. The stationary part is responsible for power generation, the moving part is in charge of the displacement of the shaft. Accordingly, the dynamic analysis of the overall action unit is derived considering the mechanical characteristics, e.g. mass, geometry, and CoM, of each part. The CoM with respect to its own frame and mass of stationary part and moving part are represented by symbols $c_{i1}$, $c_{i2}$ and $m_{i1}$, $m_{i2}$, respectively.

       \begin{eqnarray}
       \vec{c_{i1}}=A_i+c_{i1}\widehat{s_i}
        \label{eq:44}\end{eqnarray}
        \begin{eqnarray}
        \vec{c_{i2}}=A_i+{(l_i-c}_{i2})\widehat{s_i}
        \label{eq:45}\end{eqnarray}

The velocities of the stationary and moving parts are obtained by the time derivative of Eq~(\ref{eq:44}) and Eq~(\ref{eq:45}) and expressed in Eq~(\ref{eq:46}) and Eq~(\ref{eq:47}).
       \begin{eqnarray}
        \vec{v_{i1}}=c_{i1}\left(w_i\times\widehat{s_i}\ \right)
        \label{eq:46}\end{eqnarray}
       \begin{eqnarray}
       \vec{v_{i2}}=\dot{l_i}\widehat{s_i}+{(l_i-c}_{i2})\left(w_i\times\widehat{s_i}\right)
        \label{eq:47}\end{eqnarray}

        The skew-symmetric matrix form of a vector is represented by the symbol $S(.)$. Accordingly, Eqs~(\ref{eq:42}, \ref{eq:46} and \ref{eq:47}) are reconstructed in the form of $S(.)$ and expressed in Eqs~(\ref{eq:51}, \ref{eq:52}, \ref{eq:54} and \ref{eq:55}).

        \begin{eqnarray}
          w_i=\frac{S(\widehat{s_i})\dot{\delta_i}}{l_i}
        \label{eq:51}\end{eqnarray}
        \begin{eqnarray}
        \dot{w_i}=\frac{S(\widehat{s_i})\dot{\delta_i}-2\dot{l_iw_i}}{l_i}\
        \label{eq:52}\end{eqnarray}

        \begin{eqnarray}
       \vec{v_{i1}}=\frac{{-c}_{i1}}{l_i}(S(\widehat{s_i}{)^2 \dot{\delta}}_i)
        \label{eq:54}\end{eqnarray}

        \begin{eqnarray}
       \vec{v_{i2}}=(\frac{{l_i-c}_{i2}}{l_i}{S(\widehat{s_i})}^2+\ \widehat{s_i}.{\widehat{s_i}}^T\ )\dot{\delta_i}
        \label{eq:55}\end{eqnarray}

    Based on the Euler-Lagrange formalism, the kinetic energy of the linear actuator is formed as in Eq~(\ref{eq:56}).
        \begin{eqnarray}
          K_i=\frac{1}{2}\dot{{\delta_i}^T}M_i\dot{\delta_i}=\frac{1}{2}\dot{{v_{i1}}^T}m_{i1}\dot{v_{i1}}+\frac{1}{2}\dot{{v_{i2}}^T}m_{i2}\dot{v_{i2}}+\frac{1}{2}\dot{{w_i}^T}{(I}_{c_{i1}}+I_{c_{i2}})\dot{w_i}\label{eq:56}\end{eqnarray}

where $I_{c_{i1}}$ and $I_{c_{i2}}$ represent the moment of inertia (MoI) of the stationary and moving part, respectively.

    Rearranging the Eq~(\ref{eq:56}) yields the kinetic energy of the actuation unit expressed in Eq~(\ref{eq:57}).

        \begin{eqnarray}
          K_i=\frac{\dot{{\delta_i}^T}}{2}[(m_{i1}{c_{i1}}^2+m_{i2}(l_i-c_{i2})^2){S(\widehat{s_i})}^2+m_{i2}\widehat{s_i}{\widehat{s_i}}^T-\frac{S(\widehat{s_i}){(I}_{c_{i1}}+I_{c_{i2}})S(\widehat{s_i})}{{l_i}^2}]\dot{\delta_i}\label{eq:57}\end{eqnarray}

    The mass matrix, $M_i$ is extracted from the kinetic energy Eq~(\ref{eq:57}) and expressed in Eq~(\ref{eq:58}).
         \begin{eqnarray}
         M_i=(m_{i1}{c_{i1}}^2+m_{i2}(l_i-c_{i2})^2){S(\widehat{s_i})}^2+m_{i2}\widehat{s_i}{\widehat{s_i}}^T-\frac{S(\widehat{s_i}){(I}_{c_{i1}}+I_{c_{i2}})S(\widehat{s_i})}{{l_i}^2}\label{eq:58}\end{eqnarray}

        Deriving the potential energy equation in Eq~(\ref{eq:59}) of the actuation unit is also required to generate the analysis of the dynamics.
         \begin{eqnarray}
          P_i=-g^T\left(m_{i1}c_{i1}\widehat{s_i}+m_{i2}\left(l_i-c_{i2}\right)\widehat{s_i}\right)\label{eq:59}\end{eqnarray}
         , where $g$ is gravity constant, $g= [0;0;9.8]$.

         The gravity matrix is generated as in Eq~(\ref{eq:60}) by deriving the potential energy equation with respect to the variable, $\delta_i$.
         \begin{eqnarray}
         G_i=\frac{\partial P_i}{\partial \delta_i}=\left[\frac{(m_{i1}{c_{i1}}^2+m_{i2}(l_i-c_{i2})^2){S(\widehat{s_i})}^2}{l_i}-m_{i2}\widehat{s_i}.{\widehat{s_i}}^T\right]g\label{eq:60}\end{eqnarray}.

        Centrifugal and Coriolis force matrix is derived utilizing the Euler-Lagrange formalism in Eq~(\ref{eq:61}) and detailed in Eq~(\ref{eq:62}).

        \begin{eqnarray}
         C_i\delta_i={\dot{M}}_i\dot{\delta_i}-\frac{\partial(\dot{{\delta_i}^T}M_i\dot{\delta_i})}{2\partial \delta_i}\label{eq:61}\end{eqnarray}

        \begin{eqnarray}
         C_i=\frac{-m_{i2}c_{i2}\widehat{s_i}\dot{{\delta_i}^T}{S(\widehat{s_i})}^2}{{l_i}^2}-\frac{{w_i{\widehat{s_i}}^T(I}_{c_{i1}}+I_{c_{i2}})S(\widehat{s_i})}{{l_i}^2}...\notag \end{eqnarray}
         \begin{eqnarray}+\frac{2\dot{l_i}}{{l_i}^3}((m_{i1}{c_{i1}}^2+m_{i2}(l_i-c_{i2})^2-m_{i2}l_i(l_i-c_{i2})){S(\widehat{s_i})}^2+S(\widehat{s_i}){(I}_{c_{i1}}+I_{c_{i2}})S(\widehat{s_i}))\label{eq:62}\end{eqnarray}

\item{\textbf{Dynamic Analysis of End-Effector}}

The general dynamic equation of the end-effector of the robotic platform, the upper and lower platforms, is expressed in Eq~(\ref{eq:63}).

\begin{eqnarray}M_e(X)\ddot{X}+C_e(X,\dot{X})+G_e=F_e\label{eq:63}\end{eqnarray}
, where $M_e$, $C_e$, $G_e$, and $F_e$ represent mass matrix, centrifugal and Coriolis matrix, gravity vector of the end-effector, and force exerted by the end-effector, respectively. The position and orientation of the end-effector with respect to the base frame are expressed by $X=[x;\ y;\ z;\alpha;\beta;\gamma]$. The time derivative of $X$ that constitutes the velocity of the end-effector yields $\dot{X}=[\dot{x};\ \dot{y};\ \dot{z};\dot{\alpha};\dot{\beta};\dot{\gamma}]$.

The dynamic analysis of the end-effector requires transforming the variables from the base frame to their own frame through the $R^T$ matrix.

        \begin{eqnarray}
         R^T=R_x(\alpha)^T R_y(\beta)^T R_z(\gamma)^T=\left[\begin{matrix}C_\gamma C_\beta&C_\beta S_\gamma&-S_\beta\\C_\gamma S_\alpha S_\beta-C_\alpha S_\gamma & S_\gamma S_\alpha S_\beta+C_\alpha C_\gamma& S_\alpha C_\beta\\C_\gamma C_\alpha S_\beta+S_\alpha S_\gamma & S_\gamma C_\alpha S_\beta-S_\alpha C_\gamma& C_\alpha C_\beta\\\end{matrix}\right]\label{eq:64}\end{eqnarray}

The angular velocity of the end-effector with respect to its own frame is expressed utilizing the Eq~(\ref{eq:64}).

         \begin{eqnarray}
         w=R^T \left[\begin{matrix}\alpha\\\beta\\\gamma\\\end{matrix}\right]=\left[\begin{matrix}C_\gamma C_\beta&C_\beta S_\gamma&-S_\beta\\C_\gamma S_\alpha S_\beta-C_\alpha S_\gamma & S_\gamma S_\alpha S_\beta+C_\alpha C_\gamma& S_\alpha C_\beta\\C_\gamma C_\alpha S_\beta+S_\alpha S_\gamma & S_\gamma C_\alpha S_\beta-S_\alpha C_\gamma& C_\alpha C_\beta\\\end{matrix}\right] \left[\begin{matrix}\alpha\\\beta\\\gamma\\\end{matrix}\right]\label{eq:65}\end{eqnarray}

The angular acceleration of the end-effector with respect to its own frame is derived by means of $R^T$ and its time derivative, $\dot{R^T}$, and expressed in Eq~(\ref{eq:66}).

         \begin{eqnarray}
         \dot{w}=\dot{R^T} \left[\begin{matrix}\dot{\alpha}\\\dot{\beta}\\\dot{\gamma}\\\end{matrix}\right]+R^T \left[\begin{matrix}\ddot{\alpha}\\\ddot{\beta}\\\ddot{\gamma}\\\end{matrix}\right]\label{eq:66}\end{eqnarray}

The kinetic energy equation of the end-effector to be employed in the Euler-Lagrange formalism is formed as in Eq~(\ref{eq:67}).

         \begin{eqnarray}
          K_e=\frac{1}{2}\dot{{x_e}}^TM_e\dot{x_e}=\frac{1}{2}{{v_{e}}^T}m_{e}{v_{e}}+\frac{1}{2}{w}^TI_{e}w
         \label{eq:67}\end{eqnarray}
         , where $m_e$ and $I_e$ are the mass and inertia matrix of the end-effector. Here,  $v_e$ is the time derivative of Eq~(\ref{eq:30}).

         The mass matrix, $M_e$ is extracted from the kinetic energy Eq~(\ref{eq:67}) and expressed in
         \begin{eqnarray}
        M_e=\left[\begin{matrix}m_e I_{3x3}&0\\0&{\rm R} I_e R^T\\\end{matrix}\ \right]\label{eq:68}\end{eqnarray}

The potential energy equation in Eq~(\ref{eq:69}) of the end-effector is needed to analyze the dynamics of the end-effector.
         \begin{eqnarray}
         P_e=-g\left[\begin{matrix}-m_e X_c\\0\\0\end{matrix}\right]\label{eq:69}\end{eqnarray}

The gravity matrix is generated as in Eq~(\ref{eq:70}) by deriving the potential energy function with respect to the vector $X$.
         \begin{eqnarray}
         G_e=\frac{\partial P_e}{\partial X_c}=\left[\begin{matrix}-m_eg\\0\\\end{matrix}\right]\label{eq:70}\end{eqnarray}

Centrifugal and Coriolis force matrix is derived utilizing the Euler-Lagrange formula in Eq~(\ref{eq:71}).
         \begin{eqnarray}
         C_e=\left[\begin{matrix}0&0\\0&\dot{R} I_e R^T+R{S(w)I}_eR^T\\\end{matrix}\right]\label{eq:71}\end{eqnarray}

\item{\textbf{Dynamic Analysis of the Overall System}}

The intermediate variable $(\delta_i)$ configured by each linear actuator affects the position and orientation of the end-effector in task space in an independent manner. Considering the characteristics of the parallel manipulator, the configuration of each actuation unit affects the position and orientation of the end-effector's frame. Since the dynamic representation of the robotic platform, $(M_{all},$ $C_{all},$ and $G_{all})$, needs to be expressed in task space, the dynamic equation of each linear actuator written with respect to its own frame as expressed in Item~\ref{item:act_dyn} has to be stated in task space through the Jacobian matrix in Eq~(\ref{eq:68}).

    \begin{eqnarray}
         J_i=\left[\begin{matrix}I_{3x3}&-S(B_i)\\\end{matrix} T_{reverse}\right]\label{eq:72}\end{eqnarray}

The dynamic contributions of both the linear actuators and end-effector are added together as in Eq~\ref{eq:73}, \ref{eq:74}, and \ref{eq:75}.

    \begin{eqnarray}
         M_{all}=M_e+\sum_{i=1}^{3} J_i^T M_i J_i\label{eq:73}\end{eqnarray}
    \begin{eqnarray}
         C_{all}=C_e+\sum_{i=1}^{3} (J_i^T M_i J_i+J_i^T C_i J_i)\label{eq:74}\end{eqnarray}
    \begin{eqnarray}
         G_{all}=G_e+\sum_{i=1}^{3}J_i^T G_i\label{eq:75}\end{eqnarray}

Considering the dynamic effect of each member of the system, the generated force to actuate the end-effector in the workspace is calculated by the Eq~(\ref{eq:76}).

     \begin{eqnarray}
         M_{all}(X) \ddot{X}+C_{all}(X,\dot{X})+G_{all}=F_{all}\label{eq:76}\end{eqnarray}

\item{\textbf{Simulation of the Kinematics and Dynamics of the 3 DoF Robotic Manipulator}}

In this Section, the derivations expressed the kinematics and dynamics of the 3 DoF parallel manipulator are verified in the simulation environment. The overall flowchart of the simulation to represent forward and inverse kinematics and dynamics obtained in Section \emph{'Kinematic Analysis of the Manipulator'} and \emph{'Dynamic Analysis of the Manipulator'} is presented in Fig~\ref{fig:6}. 

\begin{figure}[h!]
\begin{center}
\includegraphics[width=12cm]{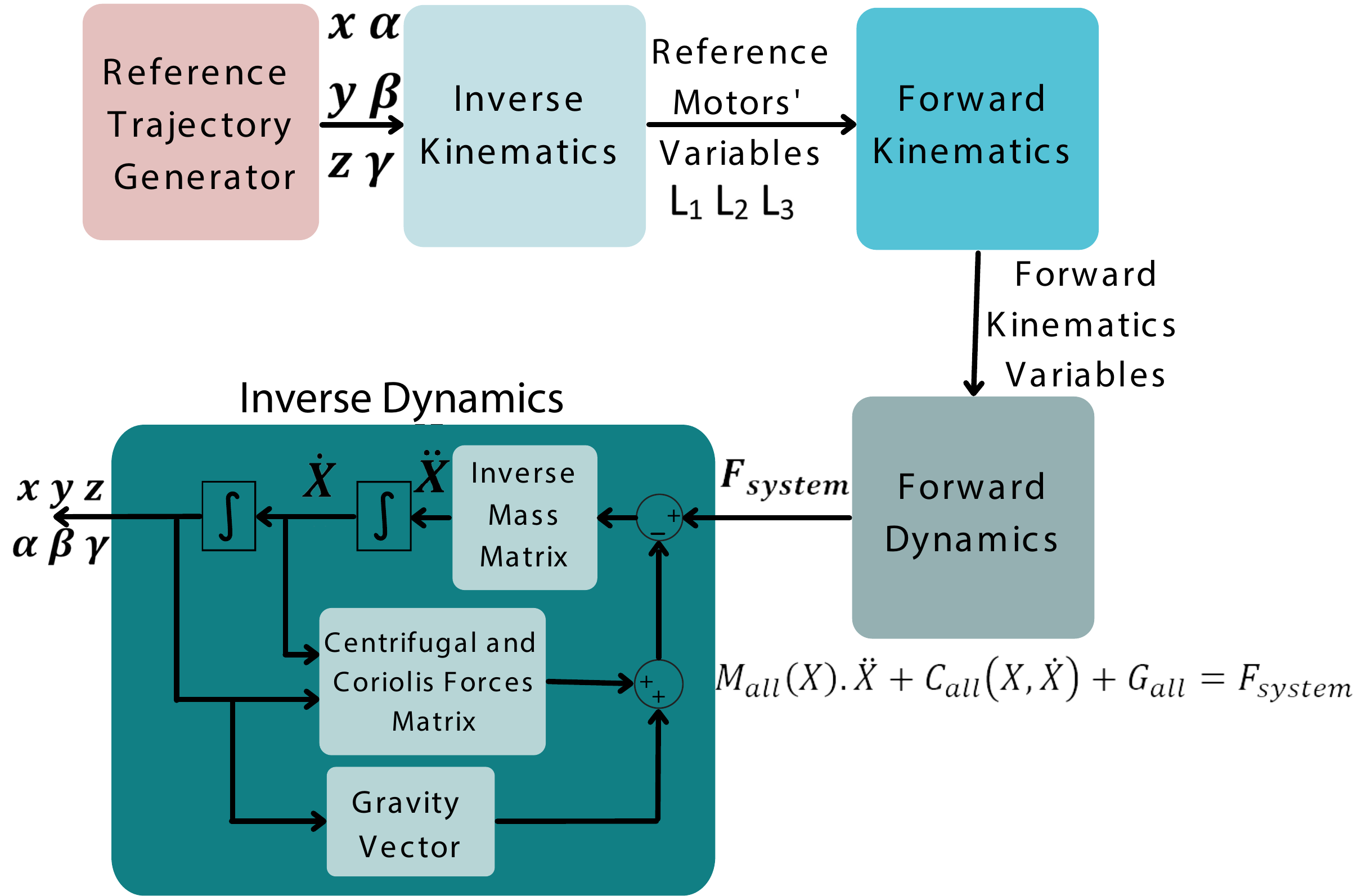}
\end{center}
\caption{The flowchart of the simulation}\label{fig:6}
\end{figure}

The dimensions of the materials produced in the first prototype, which is summarized in Table \ref{table:initialcond}, were used in the simulation. In addition, the MoI (in kg $m^2$ from the CoM of each part) of the end effector (\emph{$I_{mp}$}), moving (\emph{$I_{i1}$}) and stationary (\emph{$I_{i2}$}) parts of the linear actuator are presented as followings, respectively.

\begin{equation}
    I_{mp}=\left[\begin{matrix}0.786135&0&0\\0&1.458761&0\\0&0&0.78614\end{matrix}\right], I_{i1}=\left[\begin{matrix}0.0000525&0.0000262&0\\0.0000262&0.014331&0\\0&0&0.06169\end{matrix}\right]\notag 
\end{equation}
\begin{equation}
    I_{i2}=\left[\begin{matrix}0.003774&0&0\\0&0.000107&0\\0&0&0.003775\end{matrix}\right]\notag 
\end{equation}

The purpose of this simulation is to understand whether the kinematic and dynamic equations found are correct. The reference trajectory given at the beginning of the simulations was compared with the reference trajectory obtained as a result of inverse dynamics. In summary, the state variables, roll$(\alpha)$ and pitch$(\beta)$ angles, and $z$ translation of the end-effector (upper and lower platforms) are derived to be employed in the position, velocity, and acceleration level kinematics since only these three DoFs are controllable. Then, the constraint Eqs~(\ref{eq:13}-\ref{eq:15}) are utilized to obtain the other three dependent equalities of yaw$(\gamma)$, $x$, and $y$, and their derivatives are operated as the reference signals in velocity and acceleration levels. The reference variables defined in task space are represented in joint space with the use of inverse kinematic equations, e.g., required reference motor variables are found from reference end-effector orientation. Then, the forward kinematic representations are utilized to obtain force and torque equalities by Eqs~(\ref{eq:69}, \ref{eq:70}, and \ref{eq:71}). Lastly, the inverse dynamic analysis computes the states of the end-effector (position, velocity, and acceleration) based on torques and forces generated by the actuators.

\begin{table}[h!]
\caption{Dimensions of 3-DoF parallel manipulator, $LA^*: Linear Actuator$}
\label{table:initialcond}
\begin{tabular}{|c |c |c |c |c |c|}
 \hline
 \multirow{2}{4em}{Part} & \multirow{2}{4em}{Material} & \multirow{2}{3em}{Mass kg)} & \multirow{2}{5em}{Dimensions [$m^3$]} & \multirow{2}{6em}{Initial angle/length} & \multirow{2}{6em}{Maximum angle/length} \\
   & & & & & \\
 \hline
 \multirow{3}{4em}{Upper Platform} & ABS & 18.33 & 0.38x0.38x0.04& 0 degree & 20 degree \\
  & & & & & \\
   & & & & & \\\hline
 \multirow{3}{4em}{Lower Platform} & ABS & 14.13 & 0.38x0.38x0.03& 0 degree & 20 degree \\
   & & & & &  \\
   & & & & & \\\hline
 \multirow{3}{4em}{Stationary part of L$A^*$} & Stainless & 7.09 & 0.04x0.06x0.3 & 0.3m &- \\
   & Steel& & & & \\
   & & & & &  \\\hline
\multirow{3}{4em}{Moving part of L$A^*$} & Stainless & 1.04 & 0.2x0.02x0.02 & 0.01m & 0.2m \\ 
  &Steel& & & &  \\
   & & & & & \\\hline
\end{tabular}
\end{table}

The simulation results indicate that the actual trajectories of the end-effector in Cartesian coordinate space are well-fitted with the reference trajectories. The Fig~\ref{fig:7}A shows that the translational position of the end-effector follows the reference trajectory well with RMSE  $4.304\times10^-7$ in the x direction, $4.3\times10^-8$ in the y direction, $4.284\times10^-7$ in the z-direction. Similarly, Fig~\ref{fig:7}B provides evidence that the reference trajectory is well-matched with the rotational position of the end effector. Accordingly, the tracking errors in roll, pitch, and yaw directions are $2.824\times10^-7$, $1.316\times10^-6$, and $7.224\times10^-7$, respectively. Towards this end, the simulation results validate the accuracy of the derivations of kinematics and dynamics of the system.

\begin{figure}[h!]
\begin{center}
\includegraphics[width=13cm]{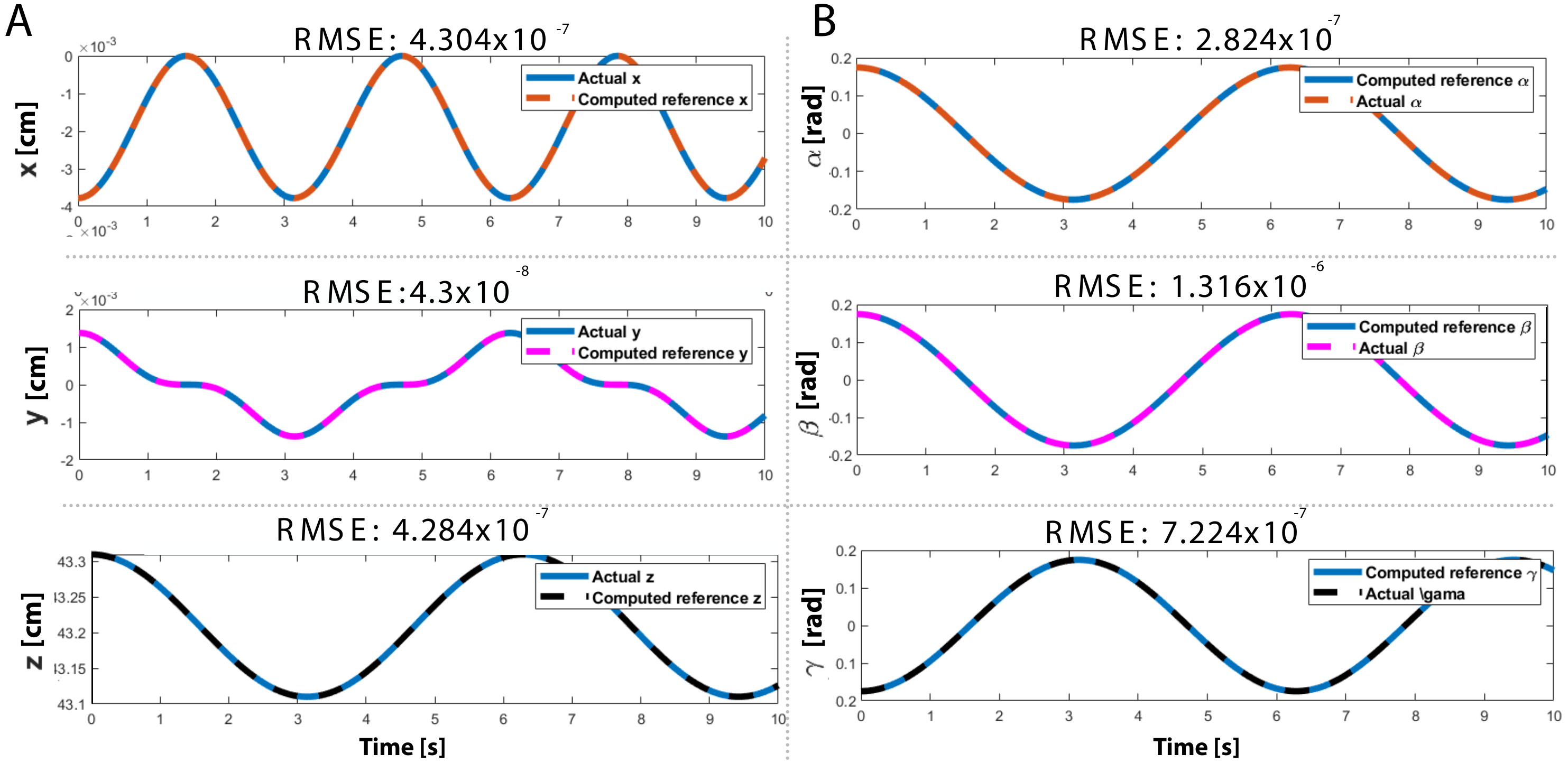}
\end{center}
\caption{Simulation results~\textbf{(A)} Translational motion of the end-effector \textbf{(B)} Rotational motion of the end-effector}\label{fig:7}
\end{figure}
\end{enumerate}

\section*{Results}
\subsection*{Stress Analyses and ADAMS simulation of the Robotic Platform}

\begin{figure}[h!]
\begin{center}
\includegraphics[width=13cm]{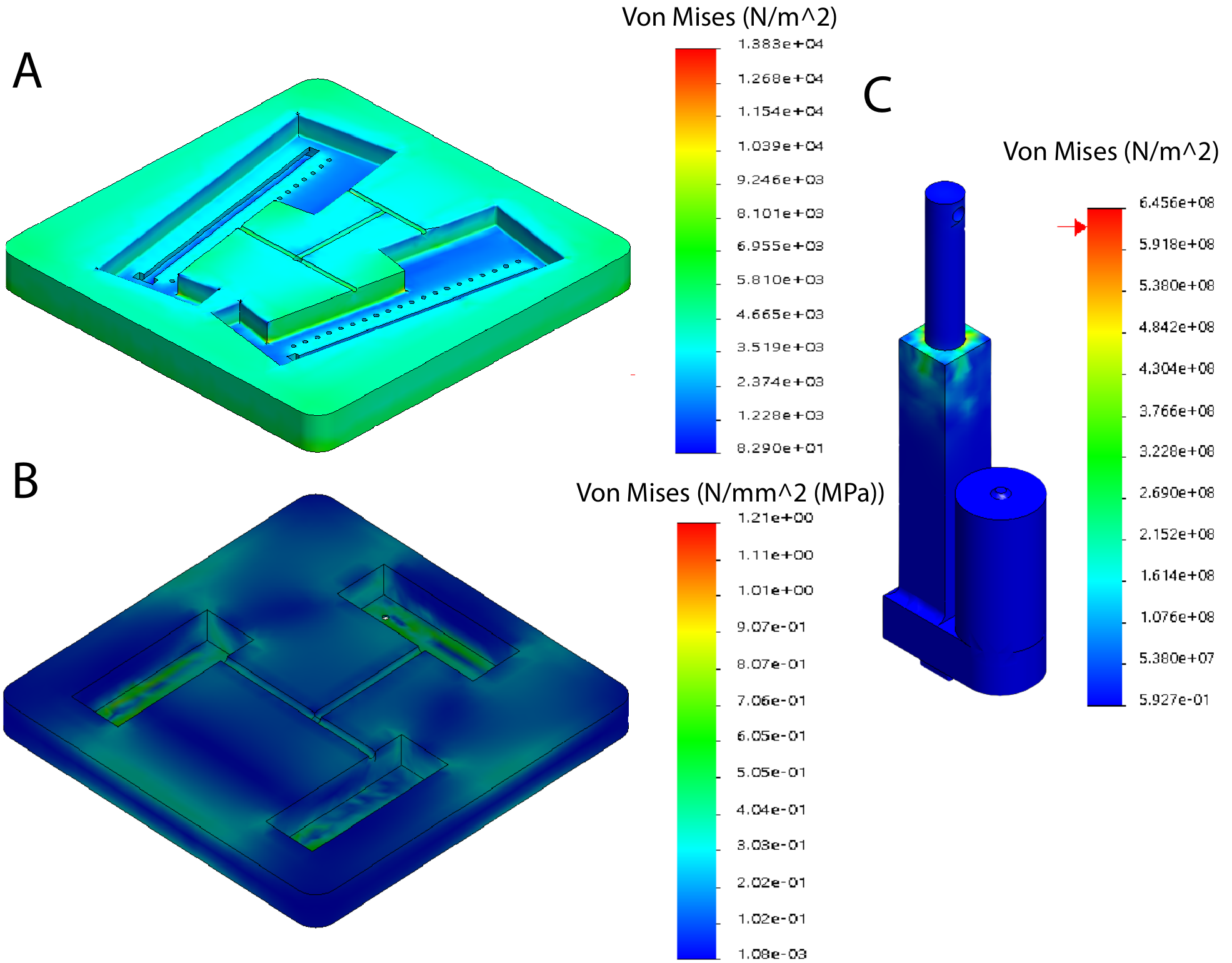}
\end{center}
\caption{Von Mises result under the static nodal stress for \textbf{(A)} Upper platform, \textbf{(B)}  Lower platform, and \textbf{(C)}  Linear actuator}\label{fig:9}
\end{figure}

The stress analyses of the robotic platform are examined as a means of finite element analysis under the loading of average human weight (687 N) and weight of the platform (120 N). In Fig~\ref{fig:9}, the response of each main part, upper and lower platforms, and linear actuators, against the static nodal stress is demonstrated. The results show that the compression force leads to less than 1mm of deformation, which is quite negligible and justifies the rigid and durable characteristics of the design. Accordingly, the analysis also supports the safety criterion of the system.

\begin{figure}[h!]
\begin{center}
\includegraphics[width=12cm]{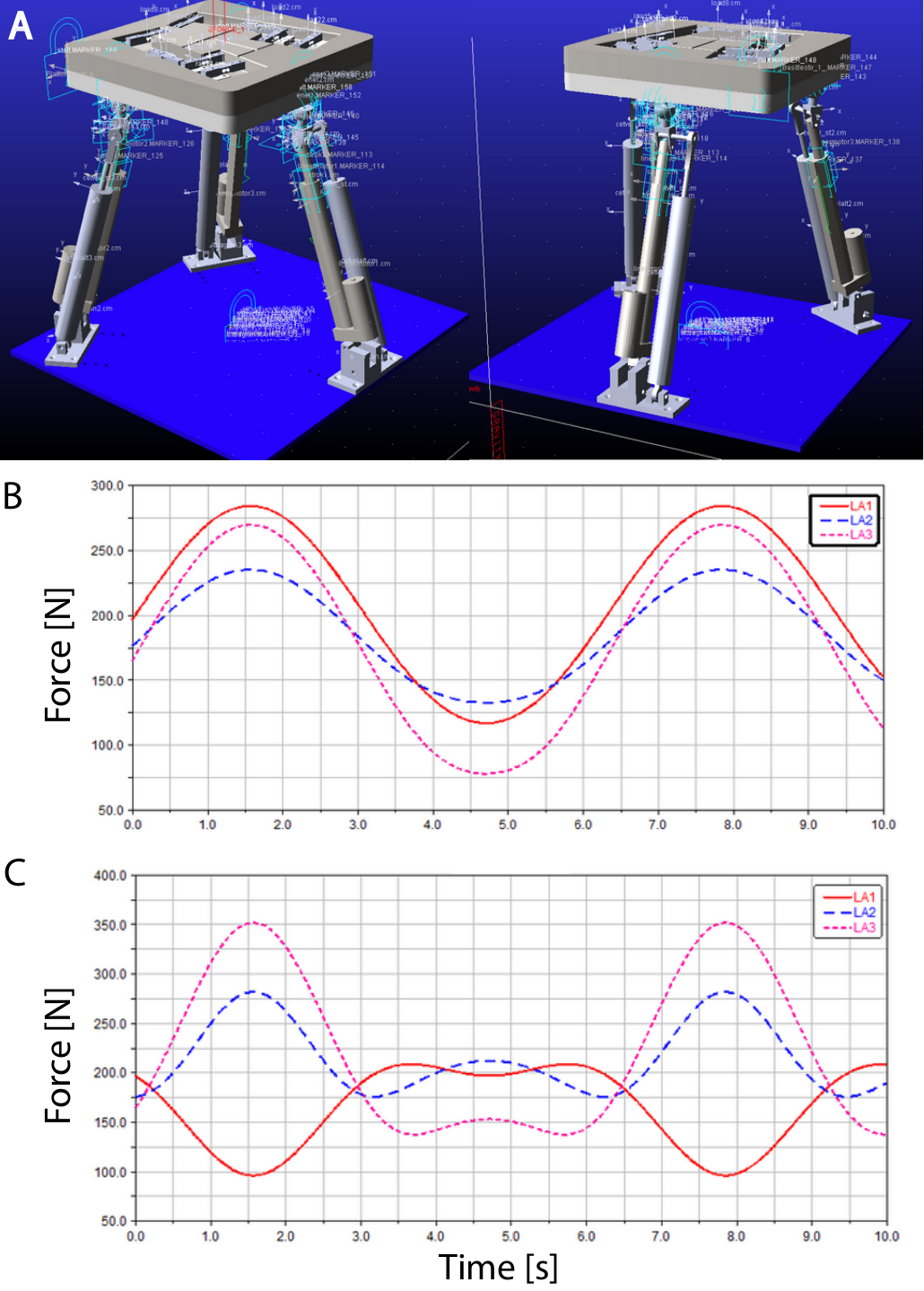}
\end{center}
\caption{ \textbf{(A)}~Adams Model, and Required linear actuator force ~\textbf{(B)}~ in z translation and ~\textbf{(C)}~ in 2$0^0$ angle}\label{fig:13}
\end{figure}

The proposed 3-DoF system was built in the ADAMS program (see Fig \ref{fig:13}A). With this simulation, the required force of the linear actuator and dynamic design simulation was performed, i.e., how each part interacts with each other, material selection, and applied force was observed in dynamic simulation. The system includes 3 RPS links, as explained in detail in Section \emph{'Kinematic and Dynamic Analyses of 3 DoF Parallel Manipulator'}. The end effector can rotate in roll $(\alpha)$ and pitch $(\beta)$ angles and translation in the z-axis or combination between these axes, i.e., each link moves with different references found by inverse kinematics equations depending on the desired end effector orientation.
The maximum human weight limit (70 kg) was applied on the end effector, and the simulation tested different orientations. When all linear actuators move in the same direction (pure z translation), the maximum force requirement is 265 N (see Fig \ref{fig:13}B). The maximum limit angle change of the platform is $20^o$ angle change, and in this case, due to all actuator tracking different reference trajectories, the maximum force, 350 N, (see Fig \ref{fig:13}C) is applied to the third linear actuator. The linear actuators have been selected to withstand about three times this power (900 N).

\subsection*{Experimental Evaluation of the System Performance}
\begin{figure}[h!]
\begin{center}
\includegraphics[width=12cm]{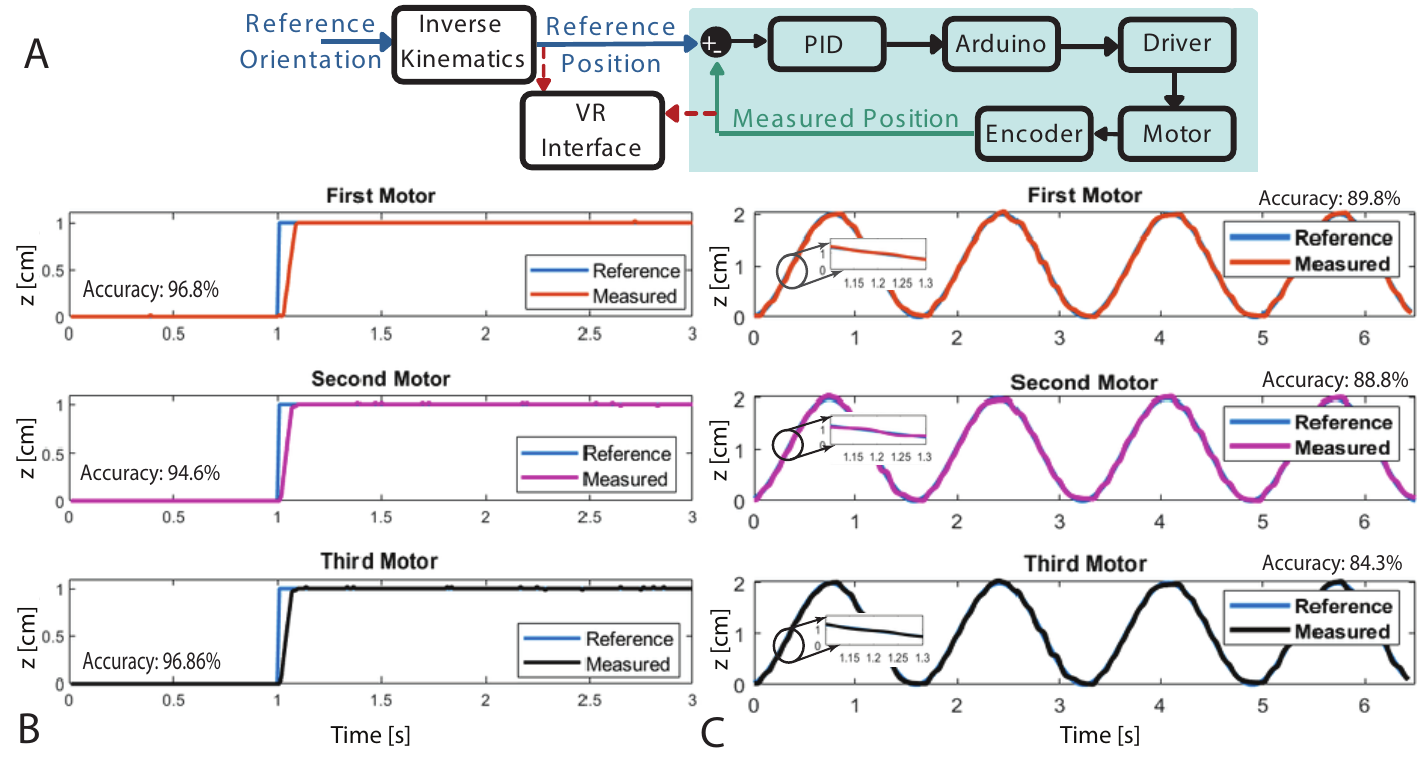}
\end{center}
\caption{Experimental verification of system performance \textbf{(A)} Real-time control architecture ~\textbf{(B)} Response of each actuator to a step reference~\textbf{(C)} Response of each actuator to a sinusoidal reference} \label{fig:10}
\end{figure}

 This Section presents the experimental evaluation of the closed-loop control of the 3 DoF planar manipulator. The aim of this experiment is to verify simulation results with the control performance of the manipulator in real-time. Three linearized DC actuators are controlled through the PID (Proportional–Integral–Derivative) controller. The actual positions of each motor are measured by linear potentiometers moving parallel to the motor shaft as shown in Fig~\ref{fig:1}. PID is employed to minimize the mismatch between the manipulator's actual performance and the reference trajectory. This real-time control architecture is implemented on MATLAB/SIMULINK environment as shown in Fig \ref{fig:10}A. PID controller tuning, the process of adjusting three parameters to fulfill stated performance standards, is realized to design the controller. In this study, the offline Grey-Box system identification method is utilized to determine the transfer function of the linear actuator as the datasheet of the actuators was not available. The output data of the model of the system presented in the open-loop control architecture was recorded by giving input signals at different frequency values. The model estimation was carried out utilizing the MATLAB System Identification Toolbox. The selected model was tuned in the MATLAB - SISO Toolbox, and the PID parameter was obtained as P: 3.66, I: 1.813, and D: 0.278e-3.  The detected control parameters were validated by simulation in MATLAB/SIMULINK environment before their use in real-time implementation.

The tracking performance of each linear actuator affects the control of the end-effector. Accordingly, the performance evaluation of each actuator is evaluated based on two different reference input signals. Fig \ref{fig:10}B presents the response of the actuator measured by the linear potentiometer when the motors are conditioned by step input. The experimental results indicate that each motor follows the given reference with tracking accuracy of $94.6\%$, $96.86\%$, and $96.8\%$. Moreover, the control performance of each motor is also evaluated based on the sine response (see Fig. \ref{fig:10}C). Accordingly, the position tracking accuracy of the first, second, and third actuators are $89.8\%$, $88.8\%$, and $84.3\%$, respectively. The small error was observed basically due to the friction in the linear potentiometer but can be inhibited by improving the mechanical imperfection inside the tube. Along these lines, the robust controllers employed with high gains and control rates allow the tracking error of the actuators at a low level.

\subsection*{Design Evaluation of the Robotic Platform}
The pressure distribution under the soles of the feet depends on the balance skill of a person while performing the stability tasks. The end-effector of the parallel manipulator is specifically designed to detect pressure distribution and the amount of CoM deviation in both static and dynamic conditions. The upper platform includes eight load cells with 10 kg load capacity in accordance with the important pressure zones~\cite{Price_2014} under the sole of the human foot, whereas, the lower platform includes three load cells with 40 kg capacity. The capacity of the load cells of the upper and lower platforms is determined based on the average pressure under the foot and the average human weight, respectively.

Along these lines, in this Section, the design evaluation of the upper and lower platforms was made by measuring the pressure applied to the load cells on the platform in different situations of the specifically prepared loads (without human-subject). Thus, the electromechanical connections and the positions of the placed load cells were tested in both static and dynamic conditions. In order to carry out these tests, water-filled jars that can mimic the weight of a human have been prepared (see Table \ref{table:1}). The prepared load weights were compared with the literature values taken from different regions of the foot's sole.

\begin{figure}[h!]
\begin{center}
\includegraphics[width=12cm]{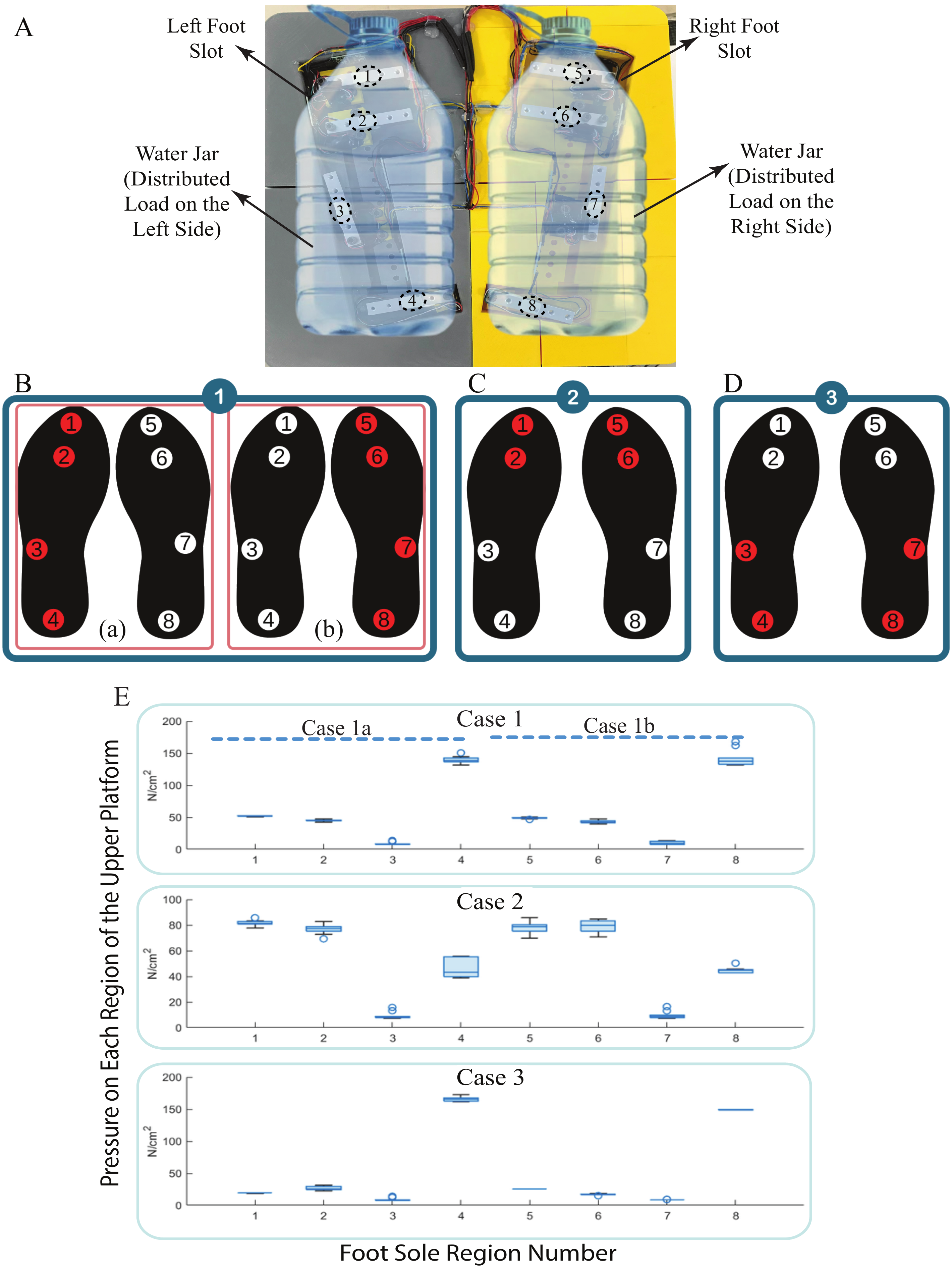}
\end{center}
\caption{Experimental procedure for the evaluation of upper platform in the static condition~\textbf{(A)}~Two water-filled jars are distributed over the upper platform~\textbf{(B)}~Pressure distribution of the distributed load on the left side of the platform only (Case 1a: standing on the left foot scenario), and right side only (Case 1b: standing on the right foot scenario)~\textbf{(C)}~Pressure distribution on the front half of the platform (Case 2: standing forefoot scenario)~\textbf{(D)}~Pressure distribution on the rear half of the platform (Case 3: standing on the rearfoot scenario)~\textbf{(E)}~Pressure distribution evaluation test statistics}\label{fig:11}
\end{figure}

First of all, the capability of static balance evaluation of the platform is tested to prove that when the system is not moving if weight is applied to a point, the load cells at that point can observe the increased pressure to different points. In this aim, three different cases of static balance/pressure distribution of a person are considered. Given the load distribution of a foot, a water-filled jar is used to transfer the load less to the toe side than to the heel side as presented in Fig~\ref{fig:11}A. In the first case, a 160 N plastic water-filled jar is placed on the left side of the upper platform to realize the standing position on the left foot (Case 1a), and only the load cell(lc)-1, lc-2, lc-3, and lc-4 are collected the data as depicted in Fig~\ref{fig:11}B. The same jar is replaced and put on the right foot slot to represent the standing position on the right foot (Case 1b), and the data due to the distributed load over the right side is gathered by the lc-5, lc-6, lc-7, and lc-8 as depicted in Fig~\ref{fig:11}B. This creates a baseline pressure distribution for the prepared load jar. In the second case, both jars are slightly tilted to the front of the platform, representing the state of inclining forward. As depicted in Fig~\ref{fig:11}C, the load distributed over the forefoot side is measured mainly by lc-1, lc-2, lc-5, and lc-6; however, other load cells also measure the remaining load on them. In the last case, the opposite scenario of Case 2 is implemented, that is, the jars are inclined to the back of the platform. Accordingly, the distributed load is mainly sensed by lc-3, lc-4, lc-7, and lc-8 as depicted in Fig~\ref{fig:11}D, but the rest of the load is also measured by other load cells.

In the static evaluation, 10 experiments were performed for each condition, each for 1 min. The distribution of the pressures of the 10 experiments for each case is given in detail in S1 Fig. The average of 10 experiments performed within each group is shown in Figure Fig~\ref{fig:11}E. The mean and standard deviation of the average collected data by each load cell is presented in Table \ref{table:3}. 

\begin{table}[h!]
\caption{Experimental evaluation of the upper platform under static condition}
\label{table:3}
\begin{tabular}{|c c c c c|}
 \hline
 Pressure [$N/cm^2$]& Case 1a & Case 1b & Case 2 & Case 3\\
 \hline
 Region 1 & 51.88±0.54 & 0  & 81.54±2.02  & 19.20±0.22 \\
\hline
 Region 2 & 44.82±1.24  & 0  & 76.77±3.94  & 26.45±3.16 \\
\hline
 Region 3 & 8.84±2.24  & 0  & 9.32±2.84  & 8.87±2.36\\
\hline
 Region 4 & 139.64±5.30  & 0  & 45.95±7.13  & 165.67±3.31 \\
 \hline
 Region 5 & 0 & 48.80±1.11   & 78.16±4.49 & 25.21±0.17 \\
 \hline
 Region 6 & 0 & 42.98±2.55  & 78.77±4.41 & 25.41±0.87 \\
 \hline
 Region 7 & 0 & 10.14±2.30 & 9.77±2.84 & 8.80±0.25\\
\hline
 Region 8 & 0 & 141.90±12.63 & 44.91±2.24 & 149.22±0.50\\
 \hline
\end{tabular}
\end{table}

\begin{figure}[h!]
\begin{center}
\includegraphics[width=12cm]{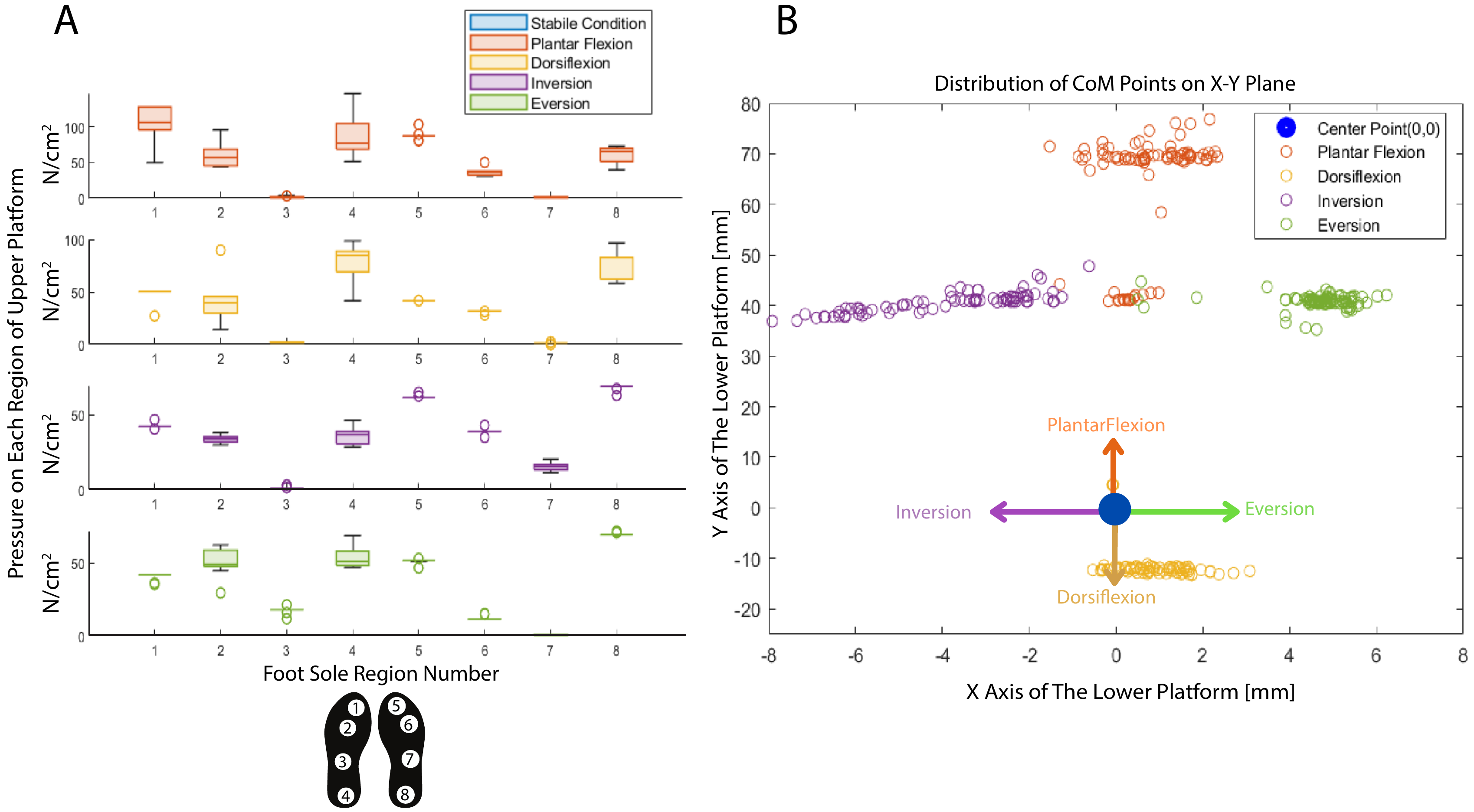}
\end{center}
\caption{Experimental evaluation of the upper and lower platforms under dynamic condition \textbf{(A)}~Pressure distribution of the payloads while the end-effector is in anthropomorphic motion ~\textbf{(B)}~Variation of CoM of payloads while the end-effector is in anthropomorphic motion}\label{fig:12}
\end{figure}

 The evaluation of the upper platform in the dynamic conditions is carried out considering the four movements of the ankle joint, namely plantar flexion, dorsiflexion, inversion, and eversion movements. In the experiment, after placing the distributed loads as shown in Fig~\ref{fig:12}A as a foot model and waiting for calibration, the manipulator follows the trajectory to realize, the standing straight condition, the plantar flexion, dorsiflexion, inversion, and eversion movements sequentially. Each movement is repeated 10 times (S2-S5 Figs) by the parallel manipulator for 1 minute, and the data gathered from each load cell is sampled at a sampling time of 1 ms. The mean and standard deviation of the collected data for each load cell is presented with box plots in Fig~\ref{fig:12}A. The two plots in Fig~\ref{fig:12}A correspond to pressure distributions of the left and right feet, respectively. The experimental results indicated in Fig~\ref{fig:12}A present that the pressure distribution due to the water-filled water jar on the platform is directly correlated with the type of movement of the platform that mimics ankle movement. That being said, the load cells located on the upper platform are sensitive to the configuration of the end effector in space. In other words, the manipulator is tilted by means of PID controller (Fig \ref{fig:10}A) as if the ankle behaves like the plantar flexion, thus the regions indicated by the numbers 1, 2 and 5, 6 of the upper platform measure higher pressure relative to the remaining parts. As for dorsiflexion type motion of the platform, lateral regions, regions 3 and 7 have higher pressures. For the inversion type motion, the load causes more pressure mainly on regions 2, 3, 6, and 7, and slightly more on regions 4 and 8, whereas for the eversion type motion of the platform, more pressure is detected on regions 2, 3, 6 and 7 and slightly more on regions 4 and 8.
 
 \begin{table}[h!]
\caption{Experimental evaluation of the upper platform under dynamic condition}
\label{table:dyn}
\begin{tabular}{|c c c c c|}
 \hline
 Pressure [$N/cm^2$]& Plantarflexion & Dorsiflexion & Inversion & Eversion\\
 \hline
 Region 1 & 102.15±29.91 & 48.15±7.36  & 40.8871±2.66  &42.50±1.63 \\
\hline
 Region 2 & 60.75±18.57  & 39.60±20.75  & 50.86±9.74  & 33.77±2.65 \\
\hline
 Region 3 & 1.96±0.75  & 1.5573±0.46  & 17.50±2.34  &1.43±0.62\\
\hline
 Region 4 & 84.41±29.08  & 79.2282±16.40  & 54.33±7.55  & 36.38±5.80 \\
 \hline
 Region 5 & 88.06±6.15 & 41.93±0.06   & 51.20±1.66 & 62.09±1.09 \\
 \hline
 Region 6 & 36.82±5.49 & 31.23±1.08  & 12.32±1.50 & 38.91±1.91 \\
 \hline
 Region 7 & 1.18±0.75 & 1.15±0.56 &0.84±0.25 & 15.49±2.81  \\
\hline
 Region 8 & 60.06±11.95 & 77.82±12.63 & 69.84±0.81 & 68.64±2.11\\
 \hline
\end{tabular}
\end{table}

 Lastly, the lower platform is also evaluated in the same experiment as both platforms are attached. During the experiment, the CoM of the water-filled jar is altered due to the different movements of the parallel manipulator as shown in Fig~\ref{fig:12}B. Each ring with a different color in Fig~\ref{fig:12}B states the CoM of the water-filled jar in an x-y coordinate system for various movements of the planar manipulator.

\section*{Discussion}
MS patients have longer reaction time, impaired balance control adjustments (APAs and CPAs), and higher fall rate and CoM deviation than their healthy pair due to affected CNS \cite{Feys_Straudi_2019, Tossavainen_2003, Cattaneo_2014, Mehravar_2015, Tijsma_2017, Aruin_Lee_2015}. In perturbation-based balance training (PBT), patients are deliberately disturbed to train their specific neuromuscular reactions, which treats the reactive balance reactions. The stability of CoM is carried out through the PBT whose characteristics allow the administration of substantial, abrupt, strong forces and force the human for successive whole-body motions. Accordingly, such a PBT program enhances the capability of patients to react to a loss of balance in unforeseen activities, and thus lower fall rates \cite{Aruin_Lee_2015, S.Aruin_2016}. In other words, the PBT program improves motor learning and develops compensatory mechanisms and strategies to overcome the loss of motor function. At the behavioral level, motor learning is considered in three phases. The first phase of learning allows the realization of the attention-demanding stage. The second phase, whereas, provides efficacy in terms of what is learned in the first phase is permanent, and accordingly alleviates unstable characteristics in motion as the error detection/correction in motor learning control architecture is minimized. Finally, movements are performed in a highly automatic and uniform manner. Thus, such training including perturbations and performed based on human feedback improves  MS patients' balance dramatically. \cite{Lee_2015, KANEKAR2015400,Hubbard_Pothier_Hughes_Rutka_2012}. Along these lines, the design objectives of the proposed device were determined within the scope of the aforementioned characteristics of motor learning.

In the literature, research on rehabilitation and assessments of MS patients has been frequently on either upper extremity~\cite{Feys_2015, Maris_2018,Gijbel_2011} or on ankle rehabilitation~\cite{5152604,Lee_Chen_Ren_Son_Cohen_Sliwa_Zhang_2017}. However, the findings in the literature show that training with the upper extremity is more beneficial to the mobility of the upper body~\cite{Feys_2015, Maris_2018}, but not effective in maintaining a balance of the body. That being said, balance loss eventuates the incidence of falls, which may lead to injury, or even death \cite{Cattaneo_2014}. Most studies designed for the lower extremity have been performed to improve the range of motion of the ankle joint \cite{Gonalves_2014,Saglia_2009,OptiFlex,Breva}. Although these studies are sufficient for the improvement of muscle weakness and instability on the ankle, balance, and motor learning are carried out by re-weighting many different conditions in standing, e.g., the entire lower extremity's strength, coordination and proprioception should be targeted at the same time \cite{Saito_2014, Hoang_2016,Hubbard_Pothier_Hughes_Rutka_2012}. However, the capacity of the end-effector of such devices is limited for single feet due to their just mere focus on ankle rehabilitation, balance rehabilitation cannot be performed with these systems. 

Today, balance rehabilitation service has been presented to patients with the use of either static or dynamic platforms. The former~\cite{Hubbard_Pothier_Hughes_Rutka_2012, Park_Lee_2014, Wii, Tyromotion} gives a general overview of the patient's balance condition~\cite{Prosperini_Pozzilli_2013,Held_Ferrer_2018}. Balance performance of the patients standing on the static platform is also evaluated by ball catching and throwing exercises \cite{Prosperini_Pozzilli_2013}. However, considering the validated efficacy of the PBT under the sole of the patients on motor learning~\cite{Held_Ferrer_2018,Hubbard_Pothier_Hughes_Rutka_2012} dynamic balance systems outperform the static ones. On the other hand, PBT is much more effective if the bio-performance of the patient is instantaneously measured and provided as feedback. Mostly, dynamic balance systems are not able to assess which side of the body is ineffective during the maintenance of the balance due to the lack of a sensor fusion system embedded end-effector. To address the problem, Hunova\cite{hunova}, 2 DoF device designed for both balance and ankle rehabilitation and able to assess sole pressure of the patients, has shown the effectiveness of the device in various clinical studies \cite{Saglia_2019,Cella_2020,Marchesi_2019,Taglione_2018,DeLuca_2020,Naro_2021}. However, the device is limited with roll and pitch rotations and thus unable to mimic the exercises provided by physiotherapists, which purposely leads to dynamic instability in the z-axis \cite{Lee_2018} such as stepping, jumping, and stair climbing exercises.

To address the above-mentioned issues, in this study we present a 3-DoF parallel manipulator designed for the lower extremity, postural control, and motor learning of MS patients. The DoF of the device covers both the ankle and balance movements in daily life activities, and it is aimed to improve the balance controls of the patients against the dynamic effects coming from the sole of the feet by giving perturbation in the z-axis. The end-effector of the parallel manipulator consists of two layers. The upper layer, called the upper platform, is equipped with eight load cells in the weight-bearing positions of the human foot. The load cells can be re-positioned for different foot sizes through the sliders mounted under the sensors. The second layer, also called the lower platform, includes three load cells positioned to form an equilateral triangle to detect the CoM of a patient. The end effector is designed to detect and train the changes in weight distribution of the body and CoM. Considering the different foot-pad pressure ratios (see Table \ref{table:1}), the pressure of each sole region is controlled based on the needs of the movement to maintain the balance. Hence, in our design, eight load cells are placed to detect the pressure ratio of each region and take a role in both the determination of the proper task based on patient-specific performance evaluation and follow-ups of the patients. In order to prove the working principle of both platforms in the end-effector, a suitable payload mimicking the force distribution transferred through the foot in both static and dynamic conditions was used in the experiments. The test executed to evaluate the payloads under static conditions was carried out in three cases as presented in Fig~\ref{fig:11}. In this experiment, first of all, the payloads employed to mimic the human feet are placed on the right and left foot regions on the upper platform to ensure that they are compatible with real human foot pad pressure. Table \ref{table:3} shows that the percentage of the load distribution is quite similar to the human foot pad pressure distribution given in Table \ref{table:1}. In the second static state, the payloads are bent towards the toes and metatarsals to mimic forefoot standing and in the third case, the loads are inclined towards the heel and middle foot. Each case is repeated 10 times, and the mean and standard deviation of the tests are summarized in Table \ref{table:3} and Fig ~\ref{fig:11}E. It has been proven that the designed payload acted like a foot pressure variation, and the upper platform detected the pressure variations in each case successfully. Accordingly, the data gathered from each load cell was provided as feedback in the control architecture of the linear actuators and VR environment, and the overall system is evaluated in the real-time dynamic environment (see Fig \ref{fig:10}). As for the experimental evaluation of the platform under dynamic conditions in a real-time environment, considering the realistic exercises directed by a physiotherapist to patients, the end-effector is forced to follow plantarflexion, dorsiflexion, inversion, and eversion orientations, respectively, which are the prominent movements that occur at the ankle joint. The experiment is repeated 10 times at different times and the pressure data gathered from each region of the upper platform is presented in Fig \ref{fig:12} and Table \ref{table:dyn}. To follow the desired trajectory, the required position reference for each actuator is obtained through the inverse kinematics, which is solved in Section of \emph{Kinematic Analysis of the Manipulator}, and the control of each is performed with negligibly small position error as shown in Fig \ref{fig:10}B and Fig \ref{fig:10}C. Along these lines, the response of the system during the dynamic condition (Fig \ref{fig:12}A) is compared with the response in the static condition (Fig \ref{fig:11}E). When the end effector is tilted forward to realize the plantarflexion movement, the payload on the toes and metatarsals of both feet increases, and the CoM reading shifts forward (pressure both in the first and fifth regions increases). Similar behavior is observed in pressure signals measured from the heel and midfoot regions as the platform is driven to perform dorsiflexion, that is, when the end effector is tilted backward (pressure both in the fourth and eighth regions increases). Since the system is inclined to the left in inversion, the pressure in the left foot is higher than in the right, and the opposite is observed in eversion. The results show that the design of the proposed robotic platform is able to sense the pressure distribution precisely and be operated in real-time, and more importantly is suitable for the experimental evaluations of the system with human subjects. 

Besides, the kinematic and dynamic modelings of the 3 DoF robotic platform are evaluated in MATLAB/SIMULINK simulation environment based on the simulation flowchart shown in Fig \ref{fig:6}. 
The reference trajectory is obtained utilizing the derived inverse kinematics for each actuator. Considering the error values between the actual and reference trajectories, the error for pitch angle variation, the maximum RMSE among the three trajectories, was found to be 1.316$\times10^-6$. Along these lines, the results (see. Fig \ref{fig:7}) of the simulations confirm that the kinematic and dynamic derivations of the 3 DoF robotic platform were found to be accurate. 

Our final work in this paper validates the simulation in a real-time environment. In other words, the digital control of each actuator was implemented by means of a PID controller, and also the compatibility of their motion with the VR environment was tested. The results indicated that each actuator was controlled in the vicinity of 95-97\% (see Fig~\ref{fig:10}). In light of all the findings, the design criteria of the 3 DoF robotic platform were validated to prove the working principle of each subsystem and to ensure that the system is suitable for evaluating its performance on the human subject in further studies.

\section*{Conclusion}
In this study, kinematic and dynamic analyses, design, implementation, and experimental evaluation of a 3-DoF robotic platform designed for the rehabilitation of balance skills and improvement of reaction time of MS patients have been presented. Simulation results and experimental evaluations indicate that the robotic platform is faithfully adapted to ankle kinematics and constraints. The novel design of the end-effector composed of two platforms is also experimentally evaluated in both static and dynamic conditions. One can deduce from the results that the proposed design is efficient to use for the evaluation of the balance skill of a person through both upper and lower platforms responsible for pressure distribution of feet and the CoM of the body, respectively. Video demonstrating the working principles of the robotic platform and its two-fold end-effector, also an illustrative example of the platform working in harmony with a task presented in a VR environment is available at S1 video \ref{S1_Video}.

The first prototype presented in the paper is a proof of concept and was implemented to provide evidence for all defined design criteria discussed in Section of \emph{Design of 3 DoF Robotic Platform}. The experiments to evaluate the end effector was evaluated by using foot-like payloads (a water-filled jar) to present the working principle of both platforms. Such an experimental object allows us to realize the realistic scenario realized with foot geometry and density, and accordingly, the pressure distribution of the object gradually increases from toe to heel, like that of a foot.
Our future work will focus on improving the capacity of both platforms equipped with load cells to conduct extensive experiments with healthy volunteers and MS patients. Such data collected during rehabilitation provides appropriate physical therapy orientation, as the robotic platform can simultaneously evaluate the user's performance based on measurements through the upper and lower platforms, and the robot actuation force and perturbation.

\section*{Supporting information}

\paragraph*{S1 Fig.}
\label{S1_Fig}
{\bf Static Design Evaluation Tests: Pressure distribution of the load cell regions under \textbf{(A)}~Case1a: standing on the left foot scenario, \textbf{(B)}~Case1b: standing on the right foot scenario,\textbf{(C)}~Case2: standing in forefoot scenario, and \textbf{(D)}~Case3: standing on the rearfoot scenario}

\paragraph*{S2 Fig.}{\bf 
Pressure distribution of the load cell regions under plantar flexion tests}\label{SFig2}

\paragraph*{S3 Fig.}{\bf Pressure distribution of the load cell regions under dorsiflexion tests}\label{SFig3}

\paragraph*{S4 Fig.}{\bf Pressure distribution of the load cell regions under inversion tests}\label{SFig4}

\paragraph*{S5 Fig.}{\bf Pressure distribution of the load cell regions under eversion tests}\label{SFig5}

\paragraph*{S1 Video.}
\label{S1_Video}
{\bf Illustrative Representation} Video demonstrates the working principles of both the robotic platform and its two-fold end-effector, also an illustrative example of the platform working in harmony with a task presented in a VR environment. The video is also available online at https://www.youtube.com/watch?v=fmmZm8K6Szg .

\paragraph*{S1 Software.}
\label{S1_Software}
{\bf Simulation of the Kinematics and Dynamics Analysis} 
MATLAB/SIMULINK Simulation file of the Kinematics and Dynamics of the 3 DoF Robotic Manipulator.

\paragraph*{S2 Software.}
\label{S2_Software}
{\bf Experimental Evaluation of the System Performance} 
MATLAB/SIMULINK file of Experimental Evaluation of the System Performance. It includes a linear actuator control structure and a VR environment interface.

\paragraph*{S3 Software.}
\label{S3_Software}
{\bf Load Data Acquisition} 
Load cell data acquisition and calibration file.

\paragraph*{S1 Data File.}
\label{S1_DataFile}
{\bf Data of Simulations and Experiments} 
All data underlying the findings are provided in the article .mat format. This file does not contain identifying participant information.


\section*{Author Contributions}
\textbf{Conceptualization:} Elif Hocaoglu \newline
\textbf{Data curation:} Tugce Ersoy \newline
\textbf{Formal analysis:} Tugce Ersoy \newline
\textbf{Investigation:} Tugce Ersoy, Elif Hocaoglu \newline
\textbf{Methodology:} Elif Hocaoglu,Tugce Ersoy \newline
\textbf{Software:} Tugce Ersoy \newline
\textbf{Project administration:} Elif Hocaoglu \newline
\textbf{Supervision:} Elif Hocaoglu \newline
\textbf{Validation:} Tugce Ersoy \newline
\textbf{Visualization:} Elif Hocaoglu,Tugce Ersoy \newline
\textbf{Writing – original draft:} Elif Hocaoglu,Tugce Ersoy

\nolinenumbers
\bibliography{refs}
\end{document}